\documentclass[10pt,twocolumn,letterpaper]{article}

\usepackage[final,algorithms]{wacv}
\usepackage{microtype}
\usepackage{array}
\usepackage{tabularx}
\usepackage{multirow}
\usepackage{fvextra}
\usepackage{xurl}
\usepackage{textcomp}
\usepackage{tikz}
\usetikzlibrary{arrows.meta,calc,positioning}

\definecolor{RetrievalPositive}{RGB}{0,135,102}
\definecolor{RetrievalNegative}{RGB}{213,94,0}
\definecolor{RetrievalQuery}{RGB}{0,114,178}
\definecolor{RetrievalNeutral}{RGB}{70,78,90}

\newcolumntype{Y}{>{\raggedright\arraybackslash}X}
\newcolumntype{P}[1]{>{\raggedright\arraybackslash}p{#1}}
\newcommand{\methodname}{CoVRAGE}
\newcommand{\methodexpansion}{%
  \underline{Co}mposed \underline{V}ideo \underline{R}etrieval with
  \underline{A}daptive \underline{G}ated \underline{E}scalation%
}

\DeclareRobustCommand{\code}[1]{\begingroup\urlstyle{tt}\nolinkurl{#1}\endgroup}
\DeclareRobustCommand{\runid}[1]{\begingroup\scriptsize\urlstyle{tt}\nolinkurl{#1}\endgroup}
\DeclareRobustCommand{\hashid}[2]{\begingroup\tiny\urlstyle{tt}\nolinkurl{#1#2}\endgroup}

\DeclareRobustCommand{\bestscore}[1]{{%
  \fontencoding{T1}\fontfamily{ptm}\fontseries{b}\fontshape{n}\selectfont #1%
}}

\DeclareRobustCommand{\ourslabel}{{%
  \fontencoding{T1}\fontfamily{ptm}\fontseries{m}\fontshape{it}\selectfont
  \methodname{} (ours)%
}}

\DeclareRobustCommand{\unverifiedtextmark}{\textsuperscript{\ensuremath{\ast}}}
\DeclareRobustCommand{\unreviewedmark}{\textsuperscript{\ensuremath{\dagger}}}

\definecolor{wacvblue}{rgb}{0.21,0.49,0.74}
\usepackage[pagebackref,breaklinks,colorlinks,allcolors=wacvblue]{hyperref}
\def\wacvPaperID{3328}
\def\confName{WACV}
\def\confYear{2027}

\title{Beyond Similarity: Foundation Models as an Efficient Backbone \\
for Training-Free Composed Video Retrieval}
\hypersetup{
  pdftitle={Beyond Similarity: Foundation Models as an Efficient Backbone for Training-Free Composed Video Retrieval},
  pdfauthor={Demidov D.},
  pdfsubject={Training-free composed video retrieval}
}
\author{
\normalsize Dmitry Demidov \\
{\tt\scriptsize dmitry.demidov@mbzuai.ac.ae}
\and
\normalsize Muhammad Zaigham Zaheer \\
{\tt\scriptsize zaigham.zaheer@mbzuai.ac.ae}
\and
\normalsize Omkar Thawakar \\
{\tt\scriptsize omkar.thawakar@mbzuai.ac.ae}
\and
\normalsize Abdelrahman Mohamed Shaker \\
{\tt\scriptsize abdelrahman.youssief@mbzuai.ac.ae}
\and
\normalsize Rao Anwer \\
{\tt\scriptsize rao.anwer@mbzuai.ac.ae}
\\
\vspace{-15.0pt}
\and
\small Mohamed bin Zayed University of Artificial Intelligence, UAE 
}

\begin{document}
\maketitle
%

\begin{abstract}
Composed video retrieval (CoVR) searches a gallery for the target video that realizes a natural-language modification of a source clip. 
However, at gallery scale, this creates a fundamental tension: compact embeddings enable efficient, reusable search but can miss the transient actions, state changes, and subtle constraints that demand fine-grained video reasoning, whereas applying large multimodal models uniformly sacrifices scalability. 
To address these limitations, we propose that frozen foundation models should instead occupy complementary roles, with inference depth adapted to query difficulty. 
Based on this premise, we introduce \methodname{}, a framework for training-free \methodexpansion{}. 
Specifically, a composed-query embedding first searches reusable video-only gallery representations; uncertain queries undergo bounded reranking and candidate expansion; ambiguous edits trigger target-description generation; and only close leading candidates reach multimodal verification. 
To support these roles, frame selection, spatial resolution, and time cues are adapted to each stage. 
Across complete target-gallery evaluations, our method reaches state-of-the-art performance among training-free approaches, with 89.55 and 93.43 R@1 on Dense-WebVid-CoVR and CoVR-R, respectively (with more than +35\% and +25\% absolute margins to the closest counterpart).
These results show that adaptively orchestrating foundation-model capabilities can combine scalable retrieval with fine-grained reasoning without task-specific training. The source code and all relevant guidelines are available on \href{https://github.com/demidovd98/CoVRAGE}{github.com/demidovd98/CoVRAGE}.

\end{abstract}

\begin{figure}[!t]
  \centering
  \includegraphics[width=1.01\columnwidth]{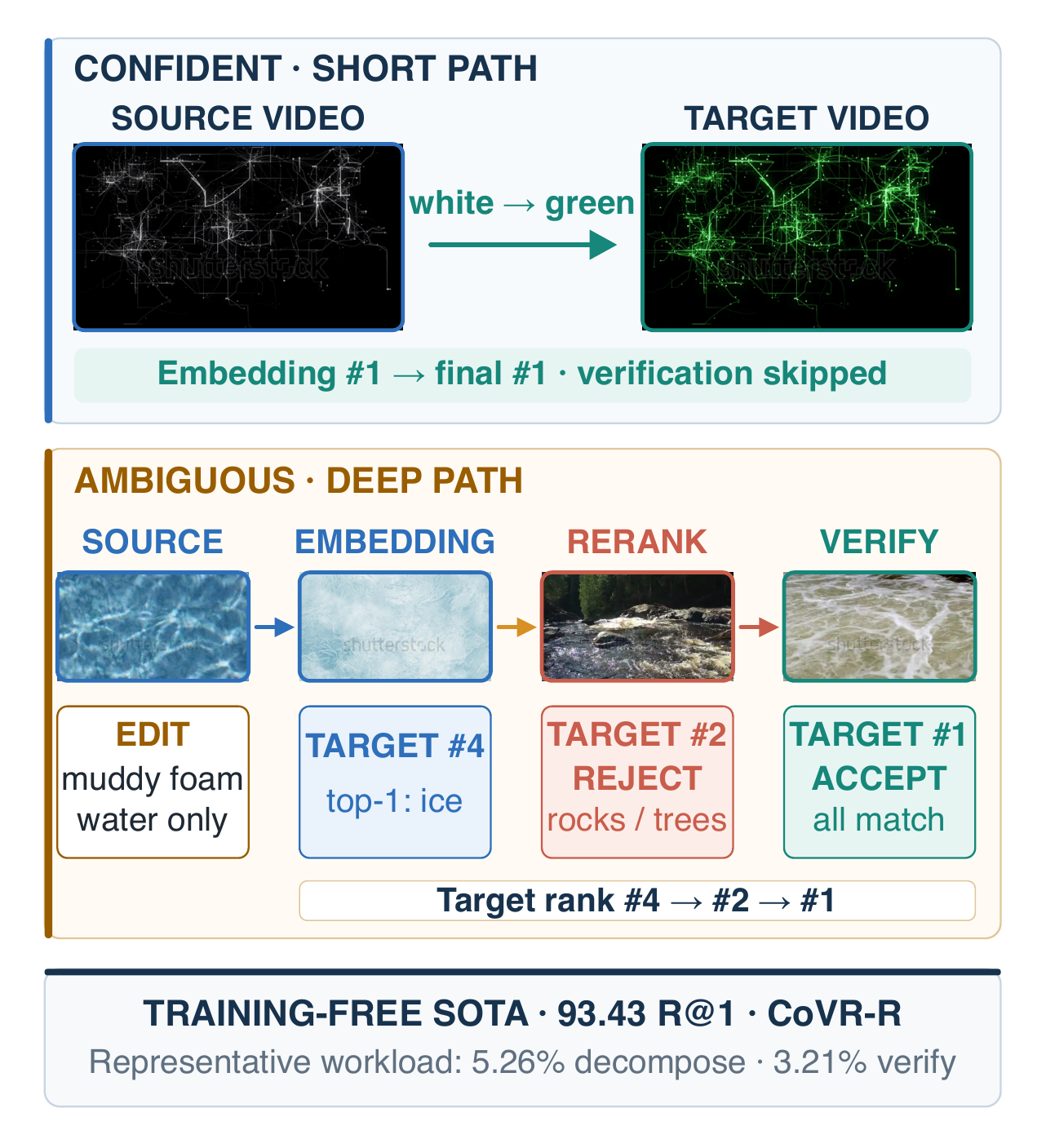}
  \vspace{-20pt}
  \caption{Two different pipeline paths through our confidence-gated cascade. A confident case skips heavy verification stages, while an ambiguous case utilizes reranking and strict verification to correct the top prediction after coarse retrieval favours a distractor. 
  The footer summarizes complete accuracy and stage-wise workload spread on the CoVR-R dataset.}
  \label{fig:first-page-teaser}
  \vspace{-5pt}
\end{figure}


\section{Introduction}\label{sec:introduction}

The composed video retrieval (CoVR) task aims to search a gallery for the desired target video that satisfies a natural-language modification to a reference clip. It offers an intuitive form of example-based search: the source specifies \emph{what to retain}, while the edit states \emph{what to change}, without requiring a complete description of the target. Composed image retrieval has progressed from learned composition of visual and textual features and free-form benchmarks to zero-shot and large-model-assisted formulations \cite{vo2019tirg,liu2021cirr,baldrati2023circo,huynh2025collm,tu2025mllmguided}, and retrieval across video and language has shown that pretrained cross-modal representations can support reusable gallery search \cite{bain2021frozen,xu2021videoclip}. Video composition, however, depends on temporal evidence: a decisive event may occupy only a few frames, and action order, duration, or a fleeting state change may distinguish otherwise similar clips. A useful system must therefore resolve fine-grained intent across time without applying expensive multi-frame reasoning to every gallery item \cite{ventura2024covr,thawakar2025dense,thawakar2026covrr}.

Recent CoVR systems address increasingly sparse, dense, egocentric, temporal, and reasoning-oriented edits through task-specific fusion, specialized objectives, generated supervision, or contextual descriptions \cite{ventura2024covr,ventura2024covr2,hummel2024egocvr,thawakar2024ecde,thawakar2025dense,thawakar2026covrr}. Frozen foundation models, target-description generation, and large-model reranking have further improved the available inference tools \cite{thawakar2026covrr,huang2026more,li2026r3}. Yet these advances expose a central tension. Large-scale pretraining yields broadly transferable representations \cite{radford2021clip,zhang2025gme}, and specialized embedders can compress a source video and edit into an indexable query for efficient comparison with reusable gallery vectors \cite{lin2025mmembed,li2026qwen3vlembedding}. Such embeddings can nevertheless lose fine-grained visual information \cite{li2025lostembeddings}; in CoVR, that bottleneck can obscure a transient action, subtle relation, or negative constraint. Applying a large reasoning model uniformly avoids this bottleneck only by multiplying computation over candidates and frames. Moreover, later reasoning can repair a ranking error only if coarse retrieval keeps the target in its candidate pool. Representation, candidate depth, and reasoning policy are therefore coupled: the representation best suited to broad search is not always the one that should make the final decision. Therefore, the resulting question is not whether one foundation model can solve CoVR, but rather how complementary capabilities should be allocated across retrieval, refinement, and reasoning.

Our premise is that frozen foundation models are most effective when assigned specialized roles in an adaptive coarse-to-fine cascade. Figure~\ref{fig:first-page-teaser} contrasts its shallow exit with its deeper corrective path. We co-design \textbf{role, routing, and reuse}: each capability addresses a particular retrieval failure, confidence determines which queries and candidates receive it, and query-independent target processing is amortized. Rather than committing every query to a fixed inference depth, the cascade treats confidence as evidence that the next capability is warranted. Confident queries terminate after compact retrieval or shallow refinement; ambiguous queries receive progressively finer candidate scoring, semantic decomposition, and multimodal verification. Stage-aware temporal evidence determines what each activated role sees. In this way, inference depth follows query difficulty: retrieve broadly, refine selectively, and reason only where ambiguity demands it.

We realize this principle with our \methodname{}, a framework for training-free \methodexpansion{}. Each gallery video is encoded once, after which a composed-query embedding retrieves a compact candidate pool. A confidence gate bypasses refinement for well-separated results; otherwise, a reranker scores the shortlist and expands it only when uncertainty persists. Terse or ambiguous edits can be converted once into a target-oriented description, and only a close leading cluster reaches strict multimodal verification. Because these stages resolve different ambiguities, their visual evidence follows the same progression: compact retrieval uses a stable uniform view, while later stages receive dynamic temporal-novelty samples, stage-specific resolution, and elapsed-time cues for irregularly spaced frames. Dynamic selection targets temporal change without increasing every frame budget, while timestamps expose elapsed time under irregular sampling. Modular interfaces separate embedding, scoring, generation, and verification. The framework thus preserves a reusable, query-independent gallery path and concentrates candidate-conditioned reasoning on a small, progressively refined set.

Across complete target-gallery evaluations, the framework reaches \textbf{54.58} R@1 on WebVid-CoVR, \textbf{89.55} on Dense-WebVid-CoVR \cite{thawakar2026covrr}, and \textbf{93.43} on CoVR-R. The latter two are the highest displayed training-free values under the evaluated public benchmark protocols, including significant absolute gains of \textbf{28.34-point R@1} and \textbf{37.95-point R@1}, respectively. Ablations show that the stages correct distinct failure modes while preserving the reusable-gallery design. Together, these findings show that adaptive allocation can combine gallery-scale retrieval with the fine-grained reasoning needed for difficult edits.

The main contributions of this research work are:
\begin{itemize}[
wide=10pt,
leftmargin=20pt
]
    \item We introduce CoVRAGE, \underline{a fully training-free, role}-\underline{specialized cascade} combining reusable gallery encoding, compact retrieval, reranking, decomposition, and verification without task-specific parameter updates.
    \item We couple confidence-controlled depth with stage-aware temporal evidence, allocating cheap candidate search, moderate visual detail refinement, and expensive reasoning according to ambiguity.
    \item We provide \underline{a controlled stage-wise study} and complete cross-dataset evaluation, while clarifying when and why each component helps.
    \item We obtain \underline{the state-of-the-art results} in training-free setup on the evaluated Dense-WebVid-CoVR \cite{thawakar2025dense} and CoVR-R \cite{thawakar2026covrr} benchmarks.
\end{itemize}

\section{Related Work}\label{sec:related-work}

\subsection{Composed Image and Video Retrieval}\label{sec:related-composed-retrieval}

Composed image retrieval represents a target from a reference image and a language edit. TIRG established explicit composition of visual and textual features \cite{vo2019tirg}; CIRR introduced natural images and free-form edits \cite{liu2021cirr}, and CIRCO extended evaluation to zero-shot, multi-positive retrieval \cite{baldrati2023circo}. Dual encoders for video and text independently showed that target videos can be encoded once and searched efficiently \cite{bain2021frozen}. CoVR joined these lines by retrieving a target video from a source visual and an edit, initially with task-trained BLIP-family composition and automatically constructed triplets \cite{ventura2024covr,ventura2024covr2}. Video adds a systems problem: decisive evidence may occupy few frames, temporal order can change the event, and query-dependent comparisons multiply multi-frame inference. Egocentric, reason-aware, and omni-modal variants further emphasize localized, implicit, or cross-modal changes \cite{hummel2024egocvr,thawakar2026covrr,ji2026omnicvr}. Thus frame allocation, temporal evidence, and candidate cost are part of the retrieval problem, not incidental implementation details.

\subsection{Supervised Composed Video Retrieval}\label{sec:related-supervised}

Direct CoVR methods have advanced through increasingly specialized supervision. Early systems fine-tune encoders and fusion modules with generated triplets, often enriching the source with generated context or dense descriptions \cite{ventura2024covr,ventura2024covr2,thawakar2024ecde,thawakar2025dense}. Later work models temporal actions, shared and differential semantics, or alignment among the source, edit, and target with task-specific losses and heads \cite{wu2025finecvr,gupta2025tfcovr,hu2026refine,zheng2026xaligner}. Other designs introduce uncertainty tokens, directional calibration, hierarchical editing, prompt modules, latent compositional schemas, audio, or rank-aware interpolation \cite{chen2025hud,li2026retrack,zhang2026relate,wang2026visionbyprompt,huang2026imagine,han2026cova,jeong2026srain}. Foundation backbones also yield strong composed embeddings after instruction fine-tuning \cite{kong2025unite}. The adapted UniCVR system learns query alignment from pseudo-triplets and conditionally deepens candidate assessment \cite{wen2026unicvr}, while interactive retrieval introduces an additional feedback loop \cite{zhang2026recovr}. These methods improve accuracy and scope, but their task-derived supervision, optimized heads, or generated training data address how to learn a CoVR model. Nevertheless, such approaches do not establish how frozen embedding, scoring, generation, and verification capabilities should be assigned and compared under one inference boundary.

\subsection{Video Retrieval with Foundational Models}\label{sec:related-foundation-models}

Instruction-tuned multimodal embedders map text, images, videos, and mixed-modal inputs into indexable representations \cite{lin2025mmembed,meng2026vlm2vecv2}. Dedicated embedding and reranking checkpoints further separate broad recall from fine-grained candidate scoring \cite{li2026qwen3vlembedding}. Universal retrieval work additionally studies modality bridging, preservation of visual identity, and any-to-any search across audio, video, and text \cite{zhang2025gme,cao2026illuminating,liu2026omniretriever}. Video-retrieval systems use these capabilities through generic video training, adaptive visual representations, or alignment to a frozen gallery \cite{halbe2026verve,gupta2026ville,wen2026unicvr}. Their supervision ranges from off-the-shelf inference to substantial adaptation, and recent diagnostics examine where video-language reasoning fails within a clip \cite{xu2026videocritic}.

The closest pipelines validate individual ingredients under different policies. CoVR-R performs generative reason-then-retrieve over reusable gallery representations \cite{thawakar2026covrr}. MoRe adds exhaustive pairwise judgments to multi-objective recall \cite{huang2026more}, while R\(^3\) combines specialized embeddings, large-model reasoning, and fixed-depth reranking \cite{li2026r3}. Parallel challenge systems explore variants that reason, retrieve, and rerank, dual-route recall, fusion of dense and sparse representations, and visual-guided video-LLM reasoning \cite{alavi2026r3covr,sun2026dualroute,liu2026reasonthenretrieve,liu2026visualguided}. These systems establish the value of reasoning, reusable recall, and candidate judgment. Our distinction is to study their allocation jointly: reusable gallery encoding, confidence-controlled depth, early and deferred decomposition, relative-cluster verification, temporal evidence, and measured activation within one fully training-free boundary.

\subsection{Efficient and Adaptive Multimodal Inference}\label{sec:related-efficient-inference}

Coarse-to-fine retrieval established the efficiency pattern of fast indexed recall followed by slower interaction over a shortlist \cite{miech2021fastslow}; specialized multimodal embedders and rerankers provide a modern instance \cite{li2026qwen3vlembedding}. Adaptive video encoders and long-video agents also vary visual computation or revisit selected evidence \cite{gupta2026ville,kurpath2026longshotbench}. Adapted staged CoVR systems add pointwise assessment after dense search \cite{wen2026unicvr}, while training-free methods generate target semantics or compare shortlisted candidates \cite{huang2026more,li2026r3}. Fixed candidate depths can nevertheless make multi-frame cost grow linearly or quadratically with pool size. Our policy instead uses independent signals: edit length launches one reusable description, the embedding gap can bypass refinement, the reranker score controls expansion, its margin can activate deferred generation, and a relative-score cluster controls verification. Query-independent novelty sampling selects the evidence seen downstream. This follows the broader principle of using embeddings for routine similarity and reserving large-model judgment for hard decisions \cite{elassadi2026embedders}.

Against this background, we provide a comprehensive stage-wise study of direct CoVR. Representation, capacity, routing, candidate depth, temporal evidence, resolution, and prompting are evaluated under one training-free pipeline and target-gallery boundary. The framework uses modular stage interfaces: reusable vectors support broad search, while confidence signals bound refinement and reserve generation and verification for ambiguity. Its contribution is the joint design and measurement of role, routing, and temporal evidence. Under the evaluated benchmark protocols, this design attains the highest training-free results on Dense-WebVid-CoVR and CoVR-R while controlling inference cost.


\section{Method}\label{sec:method}

\begin{figure*}[!t]
\centering
\includegraphics[width=\textwidth]{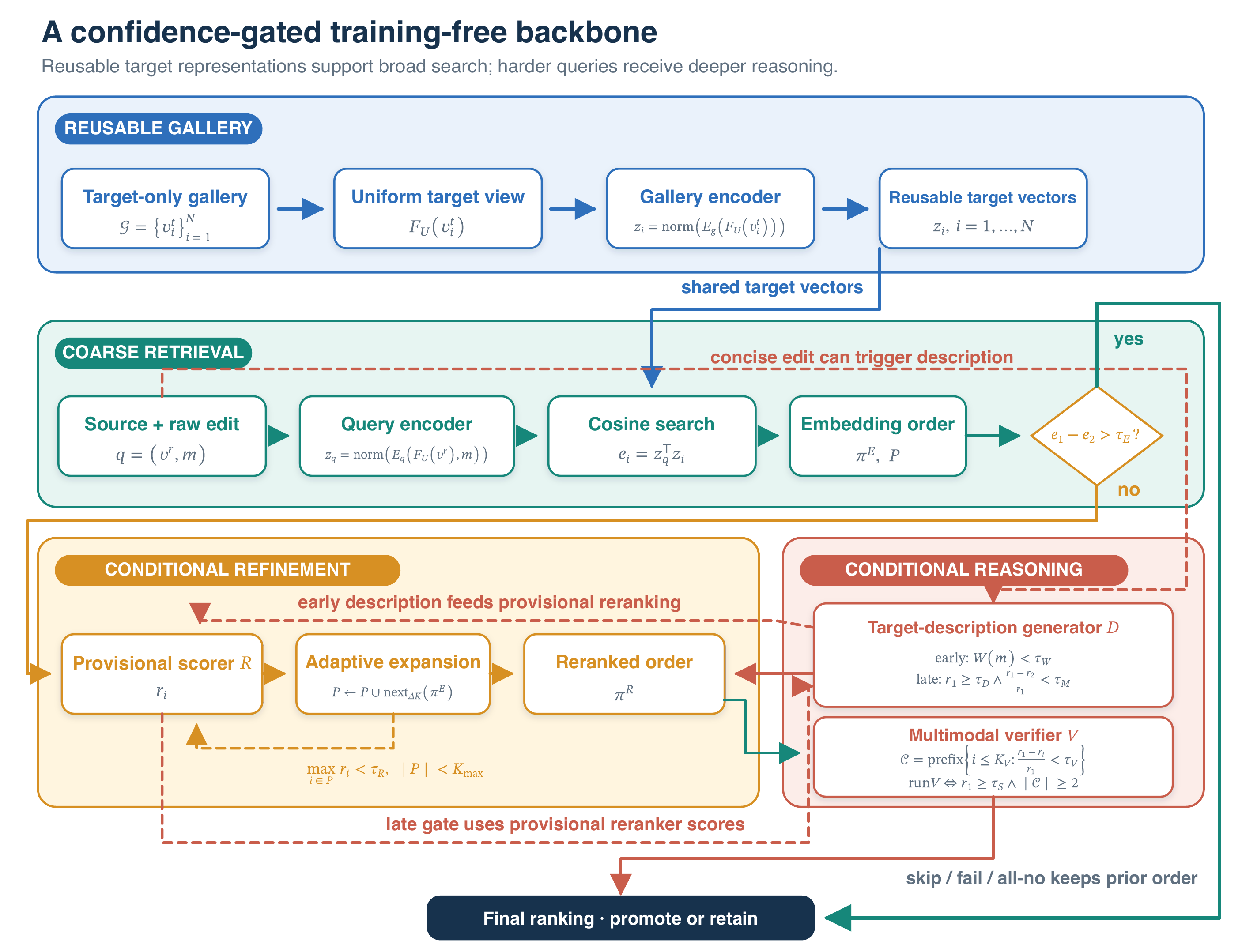}
\caption{Overview of the training-free retrieval cascade. Reusable gallery encoding supports global search, while confidence gates route only uncertain queries through bounded candidate scoring, target-description generation, and strict multimodal verification. Skipped or unresolved conditional stages retain the preceding valid ranking.}
\label{fig:model-architecture}
\end{figure*}

\subsection{Pipeline Overview}\label{sec:pipeline-overview}

Let \(v^{r}\) be a source video and \(m\) a free-form modification describing the desired change. A video gallery \(\mathcal{G}=\{v_i^{t}\}_{i=1}^{N}\), and \(v^\star\in\mathcal{G}\) is the target video that realizes \(m\) relative to \(v^{r}\). Given the composed query \(q=(v^{r},m)\), Equation~\ref{eq:retrieval-objective} defines the descending gallery order and target rank:
\begin{equation}
\label{eq:retrieval-objective}
\begin{aligned}
\pi_q
  &= \operatorname*{argsort}^{\downarrow}_{v_i^{t}\in\mathcal{G}}
     S(q,v_i^{t}), \\
\operatorname{rank}_{\pi_q}(v^\star)
  &= 1 + \sum_{v_i^{t}\ne v^\star}
     \mathbf{1}\!\left[S(q,v_i^{t})>S(q,v^\star)\right].
\end{aligned}
\end{equation}

Our goal is to place \(v^\star\) first without task-specific parameter updates. Our \methodname{} assigns complementary roles to frozen models: composed-query and gallery encoders \(E_q,E_g\), a candidate relevance scorer \(R\), a target-description generator \(D\), and a multimodal verifier \(V\). Their interfaces are modular. Gallery videos are encoded independently of any query, making their representations reusable; the cascade then transforms the source and edit into progressively more discriminative evidence over a bounded candidate set.

As illustrated in Figure~\ref{fig:model-architecture}, composed-query embedding searches the complete gallery. Confident results exit, while uncertain queries undergo bounded reranking and expansion. Conditional decomposition supplies an explicit target description, and verification compares source, edit, and candidate only for the remaining near-ties. A skipped or invalid conditional stage preserves the preceding order. Stage-aware inputs combine uniform coverage with novelty-focused evidence. The complete execution flow is detailed in App.~\ref{app:pipeline-overview}, and concrete interfaces and settings are listed in App.~\ref{app:interfaces}.


\subsection{
Query Embedding and Coarse Retrieval}\label{sec:embedding}

Coarse retrieval maps the query formed by the source and edit, as well as each target, into a shared space. The query encoder jointly represents a uniform source view and the raw modification; the gallery encoder receives only a target video under a separate neutral instruction. The query thus carries the transformation while target vectors remain reusable:
\begin{equation}
\label{eq:query-embedding}
\begin{aligned}
z_q &= \operatorname{norm}\!\left(E_q(F_U(v^{r}),m)\right),\\
z_i &= \operatorname{norm}\!\left(E_g(F_U(v_i^{t}))\right),
\qquad e_i = z_q^\top z_i .
\end{aligned}
\end{equation}

In Equation~\ref{eq:query-embedding}, \(F_U\) is uniform sampling and \(\operatorname{norm}(x)=x/\lVert x\rVert_2\). Similarities \(e_i\) define the initial order after masking a source that reappears as a non-target. Later stages receive only a small leading pool, while the complete order remains their fallback and untouched tail. Coarse retrieval therefore emphasizes gallery-wide recall; candidate-wise models handle the routed subset. Encoder interfaces, target masking, and boundary cases are detailed in App.~\ref{app:embedding}, with concrete settings in App.~\ref{app:interfaces}.


\subsection{Confidence-Gated Adaptive Reranking}\label{sec:reranking}

For uncertain queries, a specialized multimodal scorer evaluates each candidate against either the raw modification or one source-conditioned target description; the source video is absent. It returns a continuous relevance value \(r_i\).

Let \(\pi^E\) be the embedding order, \(P\) the current candidate pool, and \(\pi^R\) the resulting order. A separated embedding leader bypasses reranking. Otherwise, the initial pool expands in fixed increments only while its best candidate-wise score remains weak:
\begin{equation}
\label{eq:reranker-routing}
\begin{aligned}
e_1-e_2>\tau_E &\Rightarrow \pi^R=\pi^E,\\
e_1-e_2\le\tau_E &\Rightarrow
P\leftarrow P\cup\operatorname{next}_{\Delta K}(\pi^E),\\[-0pt]
&\hspace{5mm}\text{while }\max_{i\in P}r_i<\tau_R,\ |P|<K_{\max}.
\end{aligned}
\end{equation}

Under Equation~\ref{eq:reranker-routing}, easy queries take a shallow path, while ambiguous ones draw more candidates from the embedding order. Scored items are reordered by \(r_i\), while the unscored tail preserves a complete ranking. Provisional scores support deferred decomposition before the final text-conditioned order. Bypass or invalid evidence retains the embedding order. Full routing and fallback semantics are detailed in App.~\ref{app:reranking}, with concrete thresholds in App.~\ref{app:interfaces}.


\subsection{Conditional Query Decomposition}\label{sec:decomposition}

For short or implicit edits, decomposition generates a concise target description from the source and raw modification. This candidate-independent description replaces only the reranker's text, while embedding and verification retain the original edit.

Let \(W(m)\) measure edit length and \(r_1,r_2\) be the leading provisional reranker scores. Generation uses complementary early and late tests (one equation is broken down into three rows for readability, it reads top to bottom as one equation):
\begin{equation}
\label{eq:decomposer-routing}
\begin{gathered}
\operatorname{run}D \\
\Longleftrightarrow \\
W(m)<\tau_W
\ \lor\
\left(r_1\ge\tau_D\ \land\ \frac{r_1-r_2}{r_1}<\tau_M\right).
\end{gathered}
\end{equation}

In Equation~\ref{eq:decomposer-routing}, the first branch handles terse edits before scoring and the second handles plausible but poorly separated raw-text candidates. Deferred generation triggers one final reranking pass. Failed or empty output retains the raw edit. Exact boundaries, reuse, and scheduling are detailed in App.~\ref{app:decomposition}.


\subsection{Confidence-Gated Candidate Verification}\label{sec:verification}

For residual near-ties, a generative multimodal verifier receives the source, raw modification, and one candidate and returns a strict full-match verdict. Missing, contradicted, or uncertain requested conditions are rejected; unconstrained attributes need not match.

Verification is restricted to a contiguous leading ambiguity cluster. For reranker scores \(r_1\ge r_2\ge\cdots\),
\begin{equation}
\label{eq:verifier-cluster}
\begin{aligned}
\mathcal{C}
  &=\operatorname{prefix}\!\left\{i\le K_V:
    \frac{r_1-r_i}{r_1}<\tau_V\right\},\\
\operatorname{run}V
  &\Longleftrightarrow r_1\ge\tau_S\ \land\ |\mathcal{C}|\ge2.
\end{aligned}
\end{equation}

Under Equation~\ref{eq:verifier-cluster}, candidates in \(\mathcal{C}\) are checked top-down. The first acceptance is stably promoted and stops the scan; otherwise the reranked order remains. The score floor excludes implausible pools and the relative-gap rule excludes separated candidates. Exact gate and failure behavior are detailed in App.~\ref{app:verification}, and the prompt contract and limits are listed in App.~\ref{app:interfaces}.


\begin{figure*}[!t]
\centering
\begin{tabular}{@{}rcccccc@{}}
\scriptsize \rotatebox{90}{\hspace{7pt}Uniform} &
\includegraphics[width=.135\textwidth]{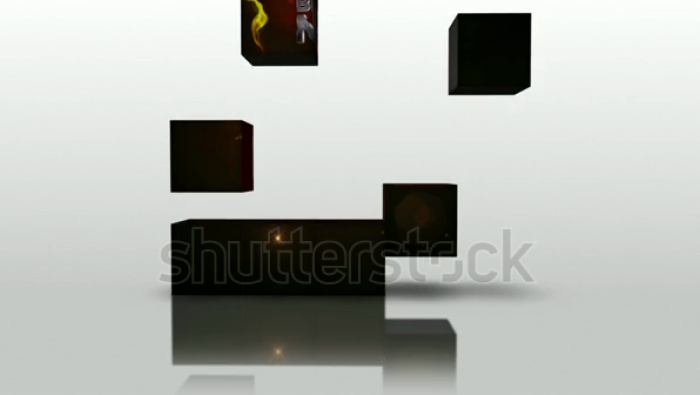} &
\includegraphics[width=.135\textwidth]{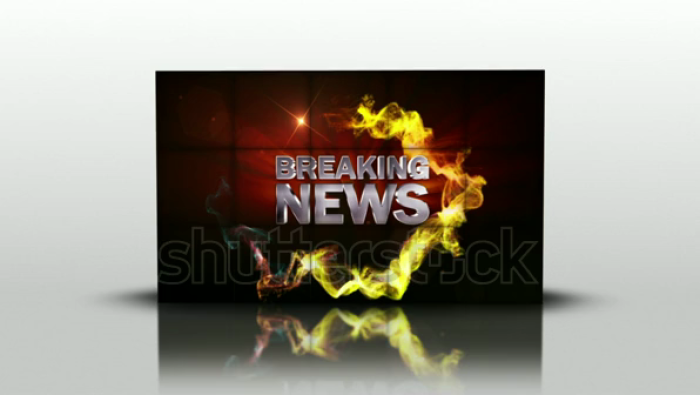} &
\includegraphics[width=.135\textwidth]{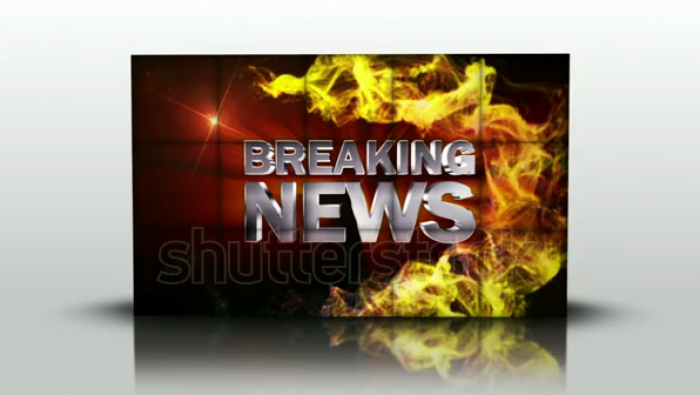} &
\includegraphics[width=.135\textwidth]{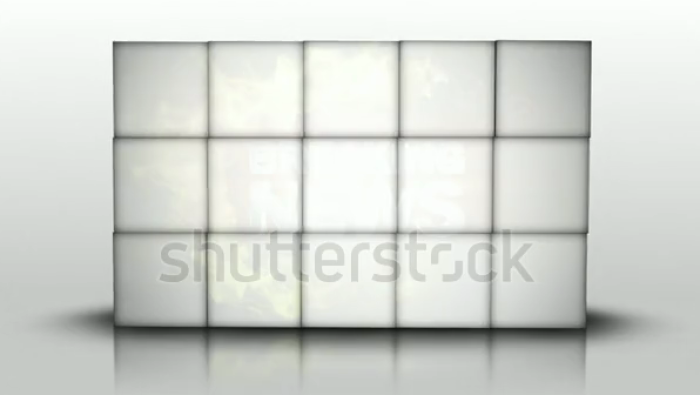} &
\includegraphics[width=.135\textwidth]{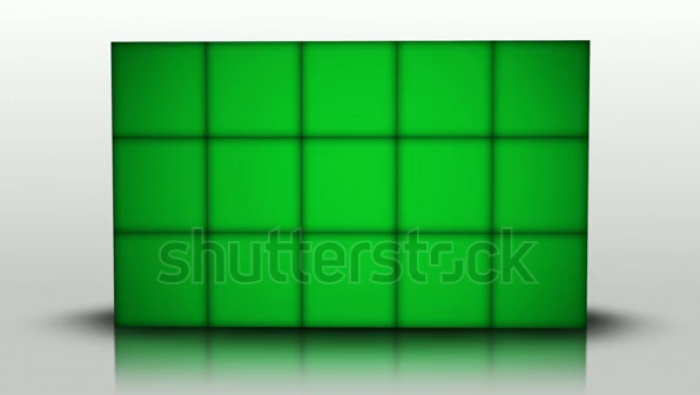} &
\includegraphics[width=.135\textwidth]{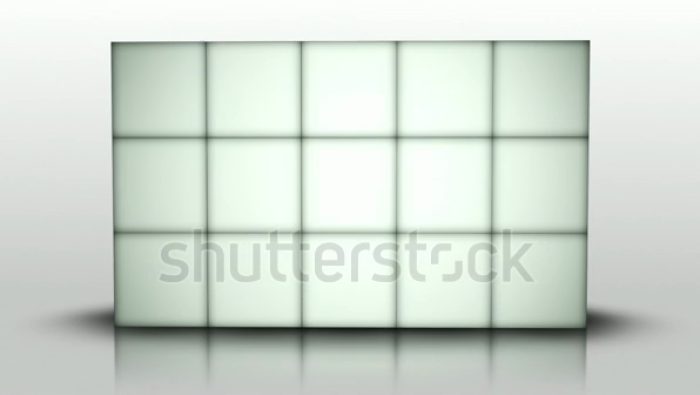} \\
\hline \\
\scriptsize \rotatebox{90}{\hspace{7pt} Dynamic} &
\includegraphics[width=.135\textwidth]{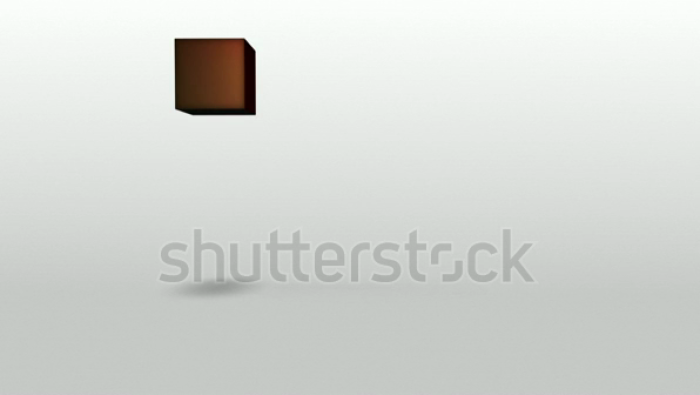} &
\includegraphics[width=.135\textwidth]{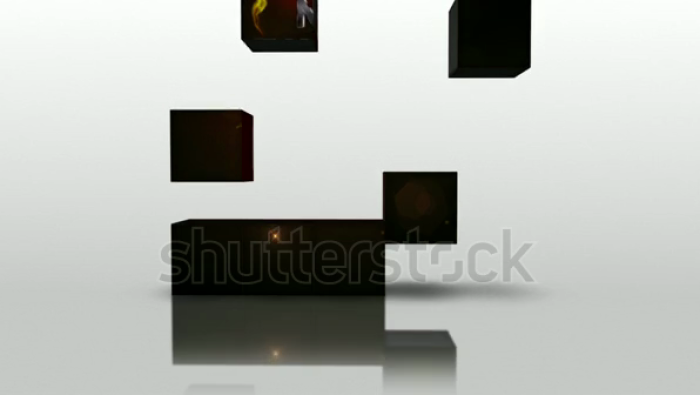} &
\includegraphics[width=.135\textwidth]{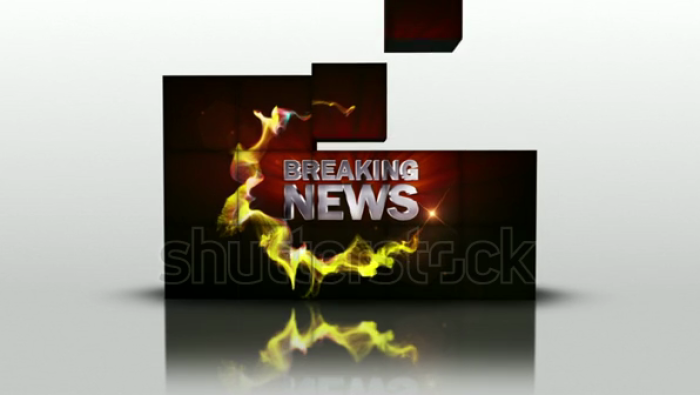} &
\includegraphics[width=.135\textwidth]{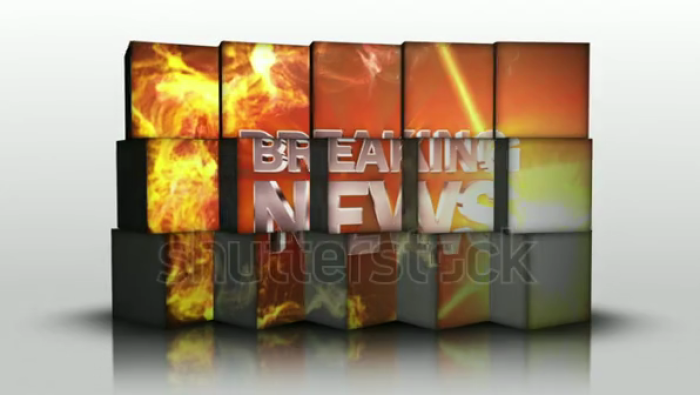} &
\includegraphics[width=.135\textwidth]{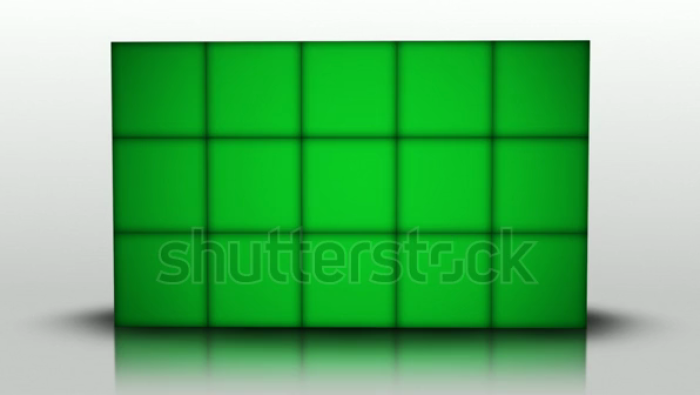} &
\includegraphics[width=.135\textwidth]{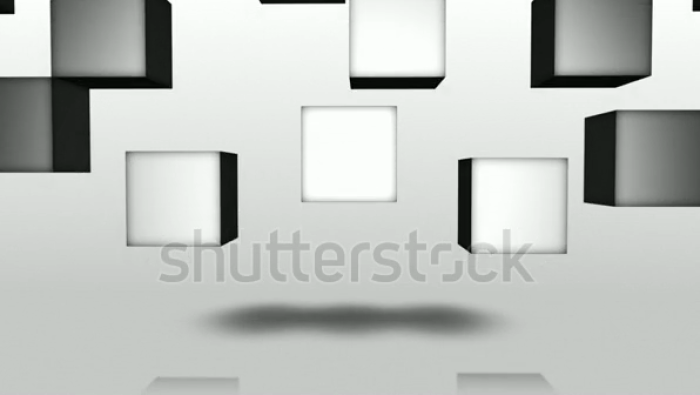}
\end{tabular}
\caption{\textbf{Dynamic frame sampling on a representative animation.} Six of the selected views are shown for each policy. Uniform spacing repeatedly samples the long title and green-screen holds; novelty weighting reallocates views to cube assembly, title reveal, the screen transition, and final disassembly. The stock watermark is retained.}
\label{fig:dynamic-sampling-showcase}
\end{figure*}

\subsection{Stage-Aware Frame Allocation}\label{sec:frame-allocation}

The cascade tailors visual evidence to each role. Global embedding uses equidistant clip views, whereas candidate scoring and generative reasoning use a query-independent novelty view derived from lightweight frame embeddings. This preserves reusable gallery indexing while focusing later stages on temporal change.

Let \(h_i\) denote the normalized embedding of temporal probe \(i\), \(\delta_i=\max(0,1-h_i^\top h_{i-1})\) its adjacent novelty, and \(B\) the downstream frame budget. The selected probe for slot \(k\) is obtained from the inverse cumulative novelty distribution,
\begin{equation}
\label{eq:dynamic-allocation}
\begin{aligned}
p_i &=
\begin{cases}
\delta_i/\sum_j\delta_j, & \sum_j\delta_j>0,\\
1/M, & \sum_j\delta_j=0,
\end{cases}\\
w_i &= \lambda p_i+\frac{1-\lambda}{M},\\[-2pt]
s_k &= \min\!\left\{i:\sum_{j\le i}w_j\ge\frac{k-\tfrac12}{B}\right\},
\quad k=1,\ldots,B.
\end{aligned}
\end{equation}

In Equation~\ref{eq:dynamic-allocation}, the explicit fallback makes \(w_i\) uniform when novelty vanishes. Inverse-CDF allocation covers the temporal support of change rather than only the largest transitions; its ordered, query-independent indices are reusable for targets. The probe schedule, duplicate handling, spatial settings, and edge cases are detailed in App.~\ref{app:frame-allocation}.

In Figure~\ref{fig:dynamic-sampling-showcase} we illustrate how novelty weighting in dynamic frame sampling reallocates views relative to naive uniform frame sampling.

\subsection{Timestamp-Aware Conditioning}\label{sec:timestamp-conditioning}

Irregular samples do not reveal elapsed time from frame order alone. The two generative stages therefore receive source-relative times identified as metadata rather than scene content, supporting ordered descriptions and cross-video temporal comparison. Embedding and reranking remain annotation-free; rendering and prompt details are given in App.~\ref{app:timestamp-conditioning}.

\section{Experiments and Analysis}\label{sec:experiments}

\subsection{Experimental Setup}\label{sec:experimental-setup}

\subsubsection{Datasets and Evaluation Protocol}\label{sec:datasets-protocol}

We evaluate on WebVid-CoVR, whose edits are short and automatically derived \cite{ventura2024covr}; Dense-WebVid-CoVR, which adds detailed attributes, actions, and temporal relations \cite{thawakar2025dense}; and CoVR-R, which combines WebVid and Something-Something-V2 examples requiring more implicit reasoning \cite{thawakar2026covrr}. All evaluations use a target-video gallery with source self-masking. We report R@1, R@5, R@10, R@50, their four-cutoff average (Avg.), and MeanR3 over R@1, R@5, and R@10 for controlled ablations. 
Table~\ref{tab:dataset-protocol} in App.~\ref{app:protocol} summarizes the evaluated splits and shared gallery/self-masking protocol, and also defines subset provenance and auxiliary metrics.

\subsubsection{Implementation Details}\label{sec:implementation-details}

The system is training-free: a compact encoder retrieves from reusable video-only gallery representations, a candidate scorer refines a confidence-dependent pool, and description generation and strict verification are invoked only for ambiguous cases. Coarse retrieval uses a uniform temporal view, whereas candidate-level and reasoning stages receive novelty-weighted frames; the generative stages also receive visible timestamps. Tables~\ref{tab:full-configuration} and~\ref{tab:prompt-interfaces} in App.~\ref{app:interfaces} list model identities, thresholds, frame and resolution settings, prompts, and generation parameters; failure handling and execution details are given in App.~\ref{app:reproducibility}.

\subsection{Quantitative Results}\label{sec:quantitative-results}

As shown in Table~\ref{tab:main-results}, our method has the highest displayed training-free R@1 on the complete Dense-WebVid-CoVR and CoVR-R evaluations, reaching 89.55 and 93.43, respectively. On WebVid-CoVR, it reaches 54.58 R@1. Prior papers use ``zero-shot'' for both frozen inference and cross-dataset transfer; we therefore separate rows by whether the reported configuration learns CoVR-specific parameters rather than by the source's section heading. WebVid-CoVR's modification candidates were machine-generated and its test examples were manually selected or filtered, but many retained edits remain short and underspecified~\cite{ventura2024covr2,thawakar2026covrr}. This annotation style may partly explain the remaining gap: sparse surface cues can reward literal similarity, while a reasoning cascade can overinterpret an ambiguous request.

\begin{table*}[!t]
\caption{Training-free approaches. 
The evaluation protocol assumes the target gallery with source self-masking. 
The Avg. column is computed from the displayed cutoffs and rounded to two decimals; missing cutoffs are not imputed. 
Within each dataset, \bestscore{boldface} and \underline{underlining} mark the best and second-best available values. \unverifiedtextmark{} marks WebVid-CoVR, which includes unverified machine-generated modification requests, which are often too short or generic~\cite{ventura2024covr2,thawakar2025dense,thawakar2026covrr}. 
CoVR-R challenge-server results using a different validation/test split are excluded.}
\label{tab:main-results}
\centering
\small
\begin{tabular}{@{}llrrrrr@{}}
\toprule
Dataset & Method & R@1 & R@5 & R@10 & R@50 & Avg. \\
\specialrule{\heavyrulewidth}{2pt}{2pt}
\multirow[t]{4}{*}{CoVR-R \cite{thawakar2026covrr}}
  & CoVR-R (Qwen3-VL-8B + reasoning)~\cite{thawakar2026covrr} & 49.88 & 66.99 & 72.97 & 85.14 & 68.75 \\
  & CoVR-R (Qwen3-VL-8B, five-round refinement)~\cite{thawakar2026covrr} & 50.56 & \underline{74.03} & \underline{81.24} & \underline{92.17} & \underline{74.50} \\
  & CoVR-R (Qwen3-VL-72B)~\cite{thawakar2026covrr} & \underline{55.48} & 72.69 & 78.57 & 87.99 & 73.68 \\
  & \ourslabel & \bestscore{93.43} & \bestscore{94.27} & \bestscore{94.61} & \bestscore{94.95} & \bestscore{94.31} \\
\specialrule{\heavyrulewidth}{3pt}{2pt}
\multirow[t]{6}{*}{Dense-WebVid-CoVR \cite{thawakar2025dense}}
  & CoVR-BLIP (Avg.)~\cite{ventura2024covr} & 38.44 & 64.96 & 71.72 & 87.12 & 65.56 \\
  & ECDE (Avg.)~\cite{thawakar2024ecde} & 40.23 & 66.38 & 74.84 & 88.12 & 67.39 \\
  & BSE-CoVR (Avg.)~\cite{thawakar2025dense} & 42.41 & 68.54 & 77.07 & 91.24 & 69.82 \\
  & CoVR-R (Qwen3-VL-8B + reasoning)~\cite{thawakar2026covrr} & \underline{61.21} & \underline{83.40} & \underline{89.39} & \bestscore{97.61} & \underline{82.90} \\
  & MoRe~\cite{huang2026more} & 49.60 & 67.00 & 77.90 & \textemdash{} & \textemdash{} \\
  & \ourslabel & \bestscore{89.55} & \bestscore{95.15} & \bestscore{96.40} & \underline{97.26} & \bestscore{94.59} \\
\specialrule{\heavyrulewidth}{3pt}{2pt}
\multirow[t]{10}{*}{WebVid-CoVR\unverifiedtextmark} \cite{ventura2024covr}
  & EgoVLPv2~\cite{pramanick2023egovlpv2} & 18.10 & 30.60 & 35.60 & \textemdash{} & \textemdash{} \\
  & LanguageBind~\cite{zhu2024languagebind} & 39.30 & 65.20 & 74.20 & \textemdash{} & \textemdash{} \\
  & CLIP (Avg.)~\cite{radford2021clip} & 44.37 & 69.13 & 77.62 & 93.00 & 71.03 \\
  & CoVR-BLIP (Avg.)~\cite{ventura2024covr} & 45.46 & 70.46 & 79.54 & 93.27 & 72.18 \\
  & CoVR-BLIP-2 (Avg.)~\cite{ventura2024covr2} & 45.66 & 71.71 & 81.30 & \underline{94.80} & \underline{73.37} \\
  & ECDE (Avg.)~\cite{thawakar2024ecde} & 47.52 & 72.18 & \underline{82.37} & \bestscore{95.06} & \bestscore{74.28} \\  
  & CoVR-R~\cite{thawakar2026covrr} & 49.15 & 70.72 & 79.25 & 93.42 & 73.14 \\
  & MoRe~\cite{huang2026more} & \bestscore{63.00} & \bestscore{83.40} & \bestscore{87.60} & \textemdash{} & \textemdash{} \\
  & TFR-CVR~\cite{hummel2024egocvr} & 51.70 & \underline{75.30} & 80.70 & \textemdash{} & \textemdash{} \\
  & \ourslabel & \underline{54.58} & 73.12 & 79.26 & 86.38 & 73.34 \\
\bottomrule
\end{tabular}
\end{table*}

\subsubsection{Cross-Dataset Performance}\label{sec:cross-dataset}

The largest displayed gaps occur on benchmarks with dense and reasoning-heavy edits: on CoVR-R, the method exceeds the highest other source-reported training-free R@1 retained in Table~\ref{tab:main-results} by 37.95 points. Against MoRe specifically, \methodname{} is substantially stronger on Dense-WebVid-CoVR: 89.55 versus 49.6 R@1, 95.15 versus 67.0 R@5, and 96.40 versus 77.9 R@10. Unlike the original WebVid-CoVR annotations, the Dense-WebVid-CoVR test modifications are detailed, fully manually verified, and corrected when needed~\cite{thawakar2025dense}, making this a more realistic evaluation of specific user intent. MoRe leads on noisy WebVid-CoVR but trails sharply on its higher-quality dense counterpart. This reversal is consistent with different sensitivity to annotation quality, although the published aggregate results do not establish its precise cause. Task-trained or adapted systems are a contextual, non-resource-equivalent comparison and are reported separately in Table~\ref{tab:task-trained-results} in App.~\ref{app:cross-dataset}; Table~\ref{tab:extended-results} there provides our full cutoff-by-cutoff metrics.

\subsubsection{Where the Gains Arise}\label{sec:stage-gains}

Candidate reranking with routed decomposition contributes most of the improvement over coarse retrieval, especially on the reasoning-heavy benchmarks. Verification supplies a smaller final correction because it only resolves the leading ambiguity cluster. R@50 is preserved by design: the reusable encoder establishes candidate coverage, while later stages reorder only a bounded head. Table~\ref{tab:stage-progression} in App.~\ref{app:stage-gains} reports the complete stage trajectories. A representative measured efficiency profile and final-configuration routing workloads are reported in App.~\ref{app:routing}.

\subsection{Qualitative Results}\label{sec:qualitative-results}

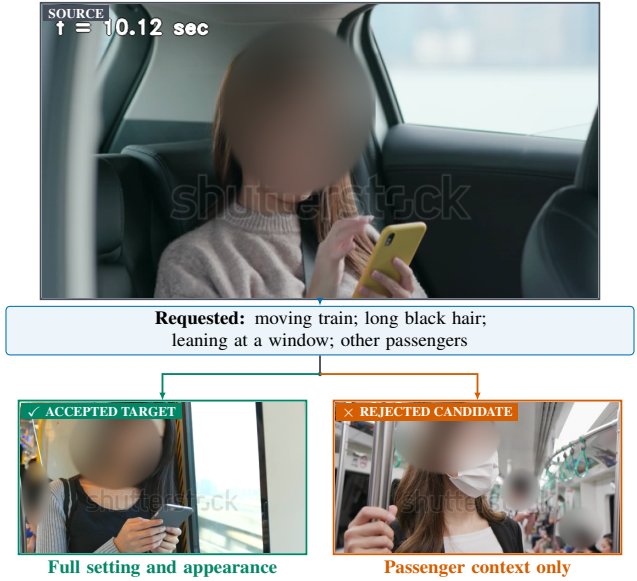
\begin{figure}[h]
\centering
\begin{tikzpicture}[
  qual image/.style={inner sep=.8pt,outer sep=0pt,line width=.7pt},
  qual pill/.style={draw=RetrievalQuery,fill=RetrievalQuery!7,
    rounded corners=1.5pt,text width=.98\columnwidth,align=center,
    inner xsep=2pt,inner ysep=2pt,font=\scriptsize},
  qual badge/.style={anchor=north west,text=white,font=\tiny\bfseries,
    inner xsep=2pt,inner ysep=1.2pt},
  qual reason/.style={anchor=north,font=\scriptsize\bfseries,align=center,
    text width=.43\columnwidth,inner sep=0pt},
  qual flow/.style={-{Latex[length=3pt,width=2.5pt]},line width=.7pt}
]
\node[qual image,draw=RetrievalNeutral] (source)
  {\includegraphics[width=.88\columnwidth]{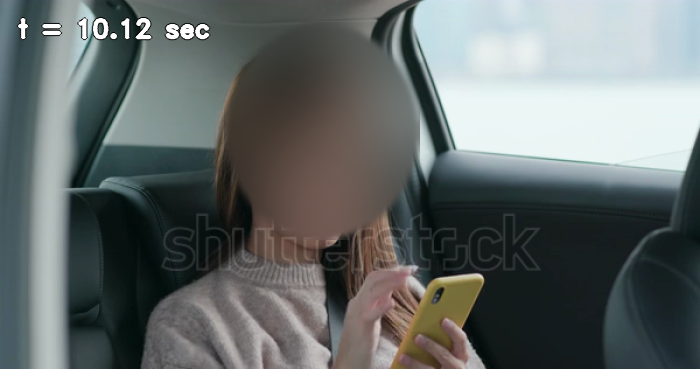}};
\node[qual pill,below=2.5pt of source] (edit)
  {\textbf{Requested:} moving train; long black hair; leaning at a window; other passengers};
\coordinate (fork) at ([yshift=-7pt]edit.south);
\node[qual image,draw=RetrievalPositive,anchor=north] (target)
  at ([xshift=-.25\columnwidth,yshift=-10pt]fork)
  {\includegraphics[width=.45\columnwidth]{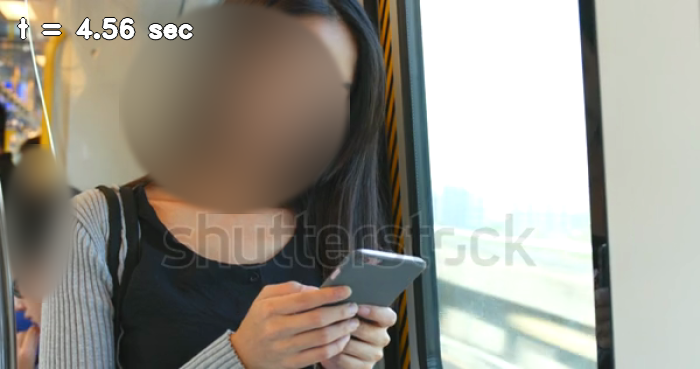}};
\node[qual image,draw=RetrievalNegative,anchor=north] (distractor)
  at ([xshift=.25\columnwidth,yshift=-10pt]fork)
  {\includegraphics[width=.45\columnwidth]{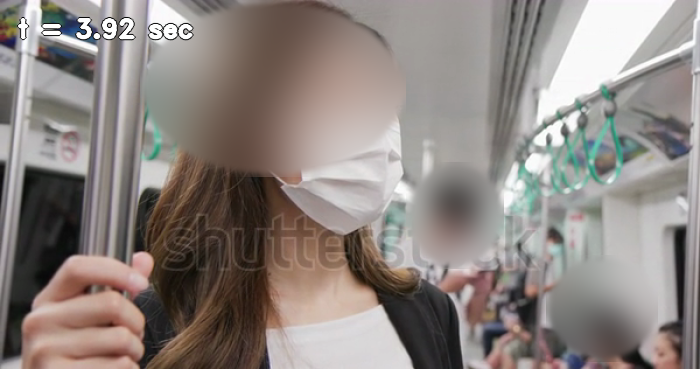}};

\draw[qual flow,RetrievalQuery] (source.south) -- (edit.north);
\draw[line width=.7pt,RetrievalNeutral] (edit.south) -- (fork);
\fill[RetrievalNeutral] (fork) circle[radius=.8pt];
\draw[qual flow,RetrievalPositive] (fork) -| (target.north);
\draw[qual flow,RetrievalNegative] (fork) -| (distractor.north);

\node[qual badge,fill=RetrievalNeutral]
  at ([xshift=.8pt,yshift=-.8pt]source.north west) {SOURCE};
\node[qual badge,fill=RetrievalPositive]
  at ([xshift=.8pt,yshift=-.8pt]target.north west)
  {$\checkmark$\;ACCEPTED TARGET};
\node[qual badge,fill=RetrievalNegative]
  at ([xshift=.8pt,yshift=-.8pt]distractor.north west)
  {$\times$\;REJECTED CANDIDATE};
\node[qual reason,text=RetrievalPositive]
  at ([yshift=-2pt]target.south) {Full setting and appearance};
\node[qual reason,text=RetrievalNegative]
  at ([yshift=-2pt]distractor.south) {Passenger context only};
\end{tikzpicture}
\caption{\textbf{Timestamp-aware verification resolves a temporal near-tie.} The upstream candidate order is unchanged, but the combined timestamp overlay and explanatory prompt let verification reject a plausible passenger clip (without leaning and phone handling in the middle) and promote the train-window target to first.}
\label{fig:qualitative-temporal-success}
\end{figure}

Qualitative behavior follows the routing policy rather than one fixed inference path. Confident queries terminate after reusable embedding or shallow refinement, whereas uncertain queries receive the particular evidence needed to resolve the remaining ambiguity. Figure~\ref{fig:qualitative-temporal-success} shows the deepest successful path: embedding and reranking preserve a plausible passenger clip ahead of the target, but timestamp-aware verification distinguishes whether the requested appearance and train-window setting are jointly satisfied.

The selected source and candidate frames are irregularly spaced and carry elapsed-time cues. Interpreted through the accompanying prompt, these cues help the verifier relate appearance and setting across each clip (with passenger picking up a phone in the middle of the video) rather than over-weighting a single passenger frame. The correction is evidence for the combined overlay-and-prompt package; it does not attribute the gain to either element in isolation.

The other routed mechanisms address earlier bottlenecks. Confidence-based expansion admits a bathtub transformation omitted from the small initial seed, after which reranking promotes it. In a controlled green-line diagnostic, target-oriented decomposition restores the missing moving-network context before retrieval and candidate scoring. These cases show that candidate depth, semantic explicitness, and temporal evidence solve complementary ambiguities. Their full stage paths are reported in App.~\ref{app:main-qualitative-relation}, and the complete qualitative casebook is provided in App.~\ref{app:retrieval-casebook}.

\subsection{Ablation Studies}\label{sec:ablations}

We retain three component-wise studies that directly explain the final cascade. All use the same Dense-WebVid-CoVR subset; further studies appear in Apps.~\ref{app:embedding-ablations} and~\ref{app:frame-sampling-ablations}.

\subsubsection{Query Strategy Analysis}\label{sec:ablation-query-strategy}

Table~\ref{tab:ablation-query-strategy} shows that joint encoding of the source video and edit improves R@1 by 14.50 points over video-only retrieval and by 4.20 over the strongest late-fusion baseline, while giving the best MeanR3 (80.73). Equal-weight late fusion is slightly stronger at R@10 (90.10 versus 89.20), and text alone is substantially weaker (36.60 R@1), supporting source-grounded composition.
\begin{table}[!h]
\caption{Embedding query strategy.
Comparing text only, source video only, equal-weight late fusion, and joint source-video/edit query encoding.
Matched training-free runs on the Dense-WebVid-CoVR subset use a target-video gallery. 
Joint encoding leads R@1 and MeanR3, supporting source-grounded composition; \textbf{bold} marks best results.
}
\label{tab:ablation-query-strategy}
\centering
\begin{tabular}{@{}lrr@{}}
\toprule
Embedding query representation & R@1 $\uparrow$ & MeanR3 $\uparrow$ \\
\midrule
Text only & 36.60 & 51.27 \\
Source video only & 52.70 & 71.13 \\
Equal-weight late fusion & 63.00 & 79.30 \\
Joint source video + edit & \textbf{67.20} & \textbf{80.73} \\
\bottomrule
\end{tabular}
\end{table}
Role-specific instructions add 3.50 R@1 in the complete sweep. We retain a joint query and reusable target vectors; full results appear in App.~\ref{app:embedding-ablations}.

\subsubsection{Score-Based Candidate Pool Expansion}\label{sec:ablation-pool-expansion}

Table~\ref{tab:ablation-pool-expansion} shows that uncertainty-triggered expansion raises R@1 from 86.10 to 89.50 and MeanR3 from 90.97 to 94.97. The full sweep in App.~\ref{app:reranker-ablations} leaves R@50 unchanged and isolates candidate depth rather than query interface. A small seed therefore suffices only for confident cases.

\begin{table}[!h]
\caption{Candidate-pool expansion for reranking.
%
Candidate-pool ablation testing whether low-confidence queries need deeper coverage than the five-item reranker seed. 
Matched training-free runs on the Dense-WebVid-CoVR subset use a target-video gallery.
Expansion improves R@1 and MeanR3, supporting confidence-gated candidate depth; \textbf{bold} marks best results.
}
\label{tab:ablation-pool-expansion}
\centering
\begin{tabular}{@{}lrr@{}}
\toprule
Reranker's pool policy & R@1 $\uparrow$ & MeanR3 $\uparrow$ \\
\midrule
Fixed initial pool & 86.10 & 90.97 \\
Confidence-based expansion & \textbf{89.50} & \textbf{94.97} \\
\bottomrule
\end{tabular}
\end{table}

\subsubsection{Text Decomposition for Reranker}\label{sec:ablation-decomposition}

Table~\ref{tab:ablation-decomposition} shows a 4.80-point R@1 gain from always-on target decomposition. Combining concise-edit and ambiguity routes retains 4.10 points while activating for 26.50\% of queries. 
\begin{table}[!h]
\caption{Text decomposition for candidate reranking.
Target-decomposition ablation testing whether source-conditioned descriptions resolve reranker ambiguity without generation for every query. 
%
%
Matched training-free runs on the Dense-WebVid-CoVR subset use a target-video gallery.
Selective routing handles edits under 10 words or ambiguous provisional scores, and Routed is the activated-query percentage. 
Always-on decomposition is most accurate, while selective routing retains 4.10 of its 4.80-point R@1 gain at 26.50\% activation, supporting the selected accuracy--workload trade-off. Best in \textbf{bold}, most optimal in \underline{underline}.
}
\label{tab:ablation-decomposition}
\centering
\begin{tabular}{@{}lrrr@{}}
\toprule
Decomposition & R@1 $\uparrow$ & MeanR3 $\uparrow$ & Routed $\downarrow$ \\
\midrule
Disabled & 89.50 & 94.97 & 0\% \\
Every query & \textbf{94.30} & \textbf{97.20} & 100\% \\
Selective & 93.60 & 96.80 & \underline{26.50\%} \\
\bottomrule
\end{tabular}
\end{table}
The larger 20.85-point WebVid-CoVR gain supports decomposition for brief edits, but embedding-side injection has mixed effects; the final interface therefore confines generated text to reranking. Routing, placement, and cross-dataset results appear in Apps.~\ref{app:decomposer-ablations} and~\ref{app:frame-sampling-ablations}.


\section{Conclusion and Discussion}\label{sec:conclusion}

We present a fully training-free CoVR cascade that combines reusable gallery search with selectively routed reranking, decomposition, and verification under stage-specific temporal evidence. It reaches \textbf{89.55 R@1} on Dense-WebVid-CoVR and \textbf{93.43 R@1} on CoVR-R, the highest results among the compared training-free methods under these protocols (Table~\ref{tab:main-results}). Qualitative and ablation evidence shows complementary corrections from temporal conditioning, joint composition, candidate depth, and semantic enrichment (Figure~\ref{fig:qualitative-temporal-success}). Apps.~\ref{app:qualitative} and~\ref{app:limitations} provide broader analysis and limitations; routing calibration and adaptive candidate budgets remain key next steps.

{\small
\bibliographystyle{ieeenat_fullname}
\bibliography{references}

\begin{thebibliography}{50}
\providecommand{\natexlab}[1]{#1}
\providecommand{\url}[1]{\texttt{#1}}
\expandafter\ifx\csname urlstyle\endcsname\relax
  \providecommand{\doi}[1]{doi: #1}\else
  \providecommand{\doi}{doi: \begingroup \urlstyle{rm}\Url}\fi

\bibitem[Alavi(2026)]{alavi2026r3covr}
Ali Alavi.
\newblock Reason, retrieve, re-rank: A zero-shot reasoning-aware framework for composed video retrieval, 2026.
\newblock arXiv preprint and CoVR-R challenge report; peer review not established.

\bibitem[Bain et~al.(2021)Bain, Nagrani, Varol, and Zisserman]{bain2021frozen}
Max Bain, Arsha Nagrani, G{\"u}l Varol, and Andrew Zisserman.
\newblock Frozen in time: A joint video and image encoder for end-to-end retrieval.
\newblock In \emph{Proceedings of the IEEE/CVF International Conference on Computer Vision}, pages 1728--1738, 2021.

\bibitem[Baldrati et~al.(2023)Baldrati, Agnolucci, Bertini, and Del~Bimbo]{baldrati2023circo}
Alberto Baldrati, Lorenzo Agnolucci, Marco Bertini, and Alberto Del~Bimbo.
\newblock Zero-shot composed image retrieval with textual inversion.
\newblock In \emph{Proceedings of the IEEE/CVF International Conference on Computer Vision}, pages 15338--15347, 2023.

\bibitem[Cao et~al.(2026)Cao, Feng, Hua, Huang, Deng, Wu, Gu, and Ye]{cao2026illuminating}
Jiawei Cao, Junyi Feng, Jiashen Hua, Ziheng Huang, Bing Deng, Kaijie Wu, Chaochen Gu, and Jieping Ye.
\newblock Illuminating visual identity in universal multimodal embeddings.
\newblock In \emph{Proceedings of the IEEE/CVF Conference on Computer Vision and Pattern Recognition}, pages 8737--8748, 2026.

\bibitem[Chen et~al.(2025)Chen, Hu, Li, Fu, Wen, and Guan]{chen2025hud}
Zhiwei Chen, Yupeng Hu, Zixu Li, Zhiheng Fu, Haokun Wen, and Weili Guan.
\newblock {HUD}: Hierarchical uncertainty-aware disambiguation network for composed video retrieval.
\newblock In \emph{Proceedings of the 33rd ACM International Conference on Multimedia}, pages 6143--6152, 2025.

\bibitem[{El Assadi} et~al.(2026){El Assadi}, Muennighoff, and Lee]{elassadi2026embedders}
Adnan {El Assadi}, Niklas Muennighoff, and Jinhyuk Lee.
\newblock The embedder's dilemma: {LLM}s are better, but at what cost?
\newblock In \emph{Third Conference on Language Modeling}, 2026.

\bibitem[Gupta et~al.(2025)Gupta, Parmar, Dave, and Shah]{gupta2025tfcovr}
Animesh Gupta, Jay Parmar, Ishan~Rajendrakumar Dave, and Mubarak Shah.
\newblock From play to replay: Composed video retrieval for temporally fine-grained videos.
\newblock In \emph{Advances in Neural Information Processing Systems: Datasets and Benchmarks Track}, 2025.

\bibitem[Gupta et~al.(2026)Gupta, Unnikrishnan, Fei, Liu, Tran, and Shah]{gupta2026ville}
Rohit Gupta, Jayakrishnan Unnikrishnan, Fan Fei, Sheng Liu, Son Tran, and Mubarak Shah.
\newblock {ViLL-E}: Video {LLM} embeddings for retrieval.
\newblock In \emph{Proceedings of the 64th Annual Meeting of the Association for Computational Linguistics (Volume 1: Long Papers)}, pages 43239--43258, San Diego, California, United States, 2026. Association for Computational Linguistics.
\newblock Outstanding Paper.

\bibitem[Halbe et~al.(2026)Halbe, Puranik, Unnikrishnan, Thakkar, Bhat, and Parag]{halbe2026verve}
Shaunak Halbe, Bhagyashree Puranik, Jayakrishnan Unnikrishnan, Kushan Thakkar, Vimal Bhat, and Toufiq Parag.
\newblock {VeRVE}: Versatile retrieval for videos via unified embeddings, 2026.
\newblock arXiv preprint arXiv:2601.12193, version 3.

\bibitem[Han et~al.(2026)Han, Jang, and Eom]{han2026cova}
Gyuwon Han, Young~Kyun Jang, and Chanho Eom.
\newblock {CoVA}: Text-guided composed video retrieval for audio-visual content.
\newblock In \emph{2026 IEEE International Conference on Acoustics, Speech and Signal Processing}, pages 12162--12166, 2026.

\bibitem[Hu et~al.(2026)Hu, Li, Chen, Huang, Fu, Xu, and Nie]{hu2026refine}
Yupeng Hu, Zixu Li, Zhiwei Chen, Qinlei Huang, Zhiheng Fu, Mingzhu Xu, and Liqiang Nie.
\newblock {REFINE}: Composed video retrieval via shared and differential semantics enhancement.
\newblock \emph{ACM Transactions on Multimedia Computing, Communications, and Applications}, 22\penalty0 (7):\penalty0 1--24, 2026.

\bibitem[Huang et~al.(2026{\natexlab{a}})Huang, Li, Chen, Fu, Wang, and Hu]{huang2026imagine}
Jiale Huang, Zixu Li, Zhiwei Chen, Zhiheng Fu, Chunxiao Wang, and Yupeng Hu.
\newblock {IMAGINE}: Adaptive schema-imagery enhanced composition for composed video retrieval.
\newblock In \emph{Proceedings of the 2026 International Conference on Multimedia Retrieval}, pages 288--297, 2026{\natexlab{a}}.

\bibitem[Huang et~al.(2026{\natexlab{b}})Huang, Wu, Jiang, Cai, Wang, and Wei]{huang2026more}
Sihong Huang, Jiaxin Wu, Dongmei Jiang, Yi Cai, Yaowei Wang, and Xiaoyong Wei.
\newblock Compositional transformation reasoning for composed video retrieval.
\newblock In \emph{Proceedings of the IEEE/CVF Conference on Computer Vision and Pattern Recognition}, pages 25644--25653, 2026{\natexlab{b}}.

\bibitem[Hummel et~al.(2024)Hummel, Karthik, Georgescu, and Akata]{hummel2024egocvr}
Thomas Hummel, Shyamgopal Karthik, Mariana-Iuliana Georgescu, and Zeynep Akata.
\newblock {EgoCVR}: An egocentric benchmark for fine-grained composed video retrieval.
\newblock In \emph{Computer Vision -- ECCV 2024}, pages 1--17, 2024.

\bibitem[Huynh et~al.(2025)Huynh, Yang, Tawari, Shah, Tran, Hamid, Chilimbi, and Shrivastava]{huynh2025collm}
Chuong Huynh, Jinyu Yang, Ashish Tawari, Mubarak Shah, Son Tran, Raffay Hamid, Trishul Chilimbi, and Abhinav Shrivastava.
\newblock {CoLLM}: A large language model for composed image retrieval.
\newblock In \emph{Proceedings of the IEEE/CVF Conference on Computer Vision and Pattern Recognition}, pages 3994--4004, 2025.

\bibitem[Jeong et~al.(2026)Jeong, Park, Kwon, Cho, and Kwak]{jeong2026srain}
Boseung Jeong, Taegyu Park, Donghyeon Kwon, Hyunsouk Cho, and Suha Kwak.
\newblock Learning sample-wise rank-aware interpolation weights for composed visual data retrieval.
\newblock In \emph{European Conference on Computer Vision}, 2026.

\bibitem[Ji et~al.(2026)Ji, Zhang, Li, Luo, Wang, Xu, Yang, Yuan, Yang, He, and Yang]{ji2026omnicvr}
Junyang Ji, Shengjun Zhang, Da Li, Yuxiao Luo, Yan Wang, Di Xu, Biao Yang, Wei Yuan, Fan Yang, Zhihai He, and Wenming Yang.
\newblock {OmniCVR}: A benchmark for omni-composed video retrieval with vision, audio, and text.
\newblock In \emph{International Conference on Learning Representations}, 2026.

\bibitem[Kong et~al.(2025)Kong, Zhang, Liu, Zhang, Feng, Yang, Wang, Tian, {Victoria W.}, Zhang, and Zhou]{kong2025unite}
Fanheng Kong, Jingyuan Zhang, Yahui Liu, Hongzhi Zhang, Shi Feng, Xiaocui Yang, Daling Wang, Yu Tian, {Victoria W.}, Fuzheng Zhang, and Guorui Zhou.
\newblock Modality curation: Building universal embeddings for advanced multimodal information retrieval, 2025.
\newblock arXiv preprint arXiv:2505.19650; no independently verified peer-reviewed acceptance at bibliography preparation time.

\bibitem[Kurpath et~al.(2026)Kurpath, Kaithakkodan, Zhou, Mullappilly, Almansoori, Ahsan, Kalmakhanbet, Shikhar, Lalla, Lahoud, Awad, Khan, Khan, Anwer, and Cholakkal]{kurpath2026longshotbench}
Mohammed~Irfan Kurpath, Jaseel~Muhammad Kaithakkodan, Jinxing Zhou, Sahal~Shaji Mullappilly, Mohammad Almansoori, Noor Ahsan, Beknur Kalmakhanbet, Sambal Shikhar, Rishabh Lalla, Jean Lahoud, Mariette Awad, Fahad~Shahbaz Khan, Salman Khan, Rao~Muhammad Anwer, and Hisham Cholakkal.
\newblock A benchmark for omni-modal reasoning in long videos, 2026.
\newblock arXiv preprint arXiv:2512.16978.

\bibitem[Li et~al.(2026{\natexlab{a}})Li, Zhang, Long, Chen, Song, Bai, Yang, Xie, Yang, Liu, Zhou, and Lin]{li2026qwen3vlembedding}
Mingxin Li, Yanzhao Zhang, Dingkun Long, Keqin Chen, Sibo Song, Shuai Bai, Zhibo Yang, Pengjun Xie, An Yang, Dayiheng Liu, Jingren Zhou, and Junyang Lin.
\newblock {Qwen3-VL-Embedding} and {Qwen3-VL-Reranker}: A unified framework for state-of-the-art multimodal retrieval and ranking, 2026{\natexlab{a}}.
\newblock arXiv preprint arXiv:2601.04720.

\bibitem[Li et~al.(2025)Li, Tang, Li, Zhang, Vuli{\'c}, and S{\o}gaard]{li2025lostembeddings}
Wenyan Li, Raphael Tang, Chengzu Li, Caiqi Zhang, Ivan Vuli{\'c}, and Anders S{\o}gaard.
\newblock Lost in embeddings: Information loss in vision--language models.
\newblock In \emph{Findings of the Association for Computational Linguistics: EMNLP 2025}, pages 22676--22693, Suzhou, China, 2025. Association for Computational Linguistics.

\bibitem[Li et~al.(2026{\natexlab{b}})Li, Hu, Chen, Huang, Qiu, Fu, and Liu]{li2026retrack}
Zixu Li, Yupeng Hu, Zhiwei Chen, Qinlei Huang, Guozhi Qiu, Zhiheng Fu, and Meng Liu.
\newblock {ReTrack}: Evidence-driven dual-stream directional anchor calibration network for composed video retrieval.
\newblock In \emph{Proceedings of the AAAI Conference on Artificial Intelligence}, pages 23373--23381, 2026{\natexlab{b}}.

\bibitem[Li et~al.(2026{\natexlab{c}})Li, Hu, Fu, Chen, Guan, and Nie]{li2026r3}
Zixu Li, Yupeng Hu, Zhiheng Fu, Zhiwei Chen, Weili Guan, and Liqiang Nie.
\newblock {$R^3$}: Composed video retrieval via reasoning-guided recalling and re-ranking, 2026{\natexlab{c}}.
\newblock arXiv preprint and CoVR-R challenge report; peer review not established.

\bibitem[Lin et~al.(2025)Lin, Lee, Shoeybi, Lin, Catanzaro, and Ping]{lin2025mmembed}
Sheng-Chieh Lin, Chankyu Lee, Mohammad Shoeybi, Jimmy Lin, Bryan Catanzaro, and Wei Ping.
\newblock {MM-Embed}: Universal multimodal retrieval with multimodal {LLM}s.
\newblock In \emph{International Conference on Learning Representations}, 2025.

\bibitem[Liu et~al.(2026{\natexlab{a}})Liu, Qi, and Ji]{liu2026reasonthenretrieve}
DongQing Liu, MengShi Qi, and HongWei Ji.
\newblock Reason-then-retrieve for {CoVR-R} with structured edit prompts and dense-sparse fusion, 2026{\natexlab{a}}.
\newblock arXiv preprint and CoVR-R challenge report; peer review not established.

\bibitem[Liu et~al.(2026{\natexlab{b}})Liu, Wu, Zhou, and Shen]{liu2026omniretriever}
Yunze Liu, Chi-Hao Wu, Enmin Zhou, and Junxiao Shen.
\newblock {OmniRetriever}: Any-to-any audio-video-text retrieval via fusion-as-teacher distillation, 2026{\natexlab{b}}.
\newblock arXiv preprint arXiv:2605.26641.

\bibitem[Liu et~al.(2026{\natexlab{c}})Liu, Xu, Wen, Dai, and Huang]{liu2026visualguided}
Yang Liu, Qianqian Xu, Peisong Wen, Siran Dai, and Qingming Huang.
\newblock Training-free composed video retrieval via visual representation-guided video-{LLM} reasoning, 2026{\natexlab{c}}.
\newblock arXiv preprint and CoVR-R challenge report; arXiv comments mention the CVPR 2026 VidLLMs workshop, but no archival proceedings record was available at bibliography preparation time.

\bibitem[Liu et~al.(2021)Liu, Rodriguez-Opazo, Teney, and Gould]{liu2021cirr}
Zheyuan Liu, Cristian Rodriguez-Opazo, Damien Teney, and Stephen Gould.
\newblock Image retrieval on real-life images with pre-trained vision-and-language models.
\newblock In \emph{Proceedings of the IEEE/CVF International Conference on Computer Vision}, pages 2125--2134, 2021.

\bibitem[Meng et~al.(2026)Meng, Jiang, Liu, Su, Yang, Fu, Qin, Thirukovalluru, Zhang, Chen, Xu, Xiong, Zhou, Chen, and Yavuz]{meng2026vlm2vecv2}
Rui Meng, Ziyan Jiang, Ye Liu, Mingyi Su, Xinyi Yang, Yuepeng Fu, Can Qin, Raghuveer Thirukovalluru, Xuan Zhang, Zeyuan Chen, Ran Xu, Caiming Xiong, Yingbo Zhou, Wenhu Chen, and Semih Yavuz.
\newblock {VLM2Vec-V2}: Advancing multimodal embedding for videos, images, and visual documents.
\newblock \emph{Transactions on Machine Learning Research}, 2026.

\bibitem[Miech et~al.(2021)Miech, Alayrac, Laptev, Sivic, and Zisserman]{miech2021fastslow}
Antoine Miech, Jean-Baptiste Alayrac, Ivan Laptev, Josef Sivic, and Andrew Zisserman.
\newblock Thinking fast and slow: Efficient text-to-visual retrieval with transformers.
\newblock In \emph{Proceedings of the IEEE/CVF Conference on Computer Vision and Pattern Recognition}, pages 9826--9836, 2021.

\bibitem[Pramanick et~al.(2023)Pramanick, Song, Nag, Lin, Shah, Shou, Chellappa, and Zhang]{pramanick2023egovlpv2}
Shraman Pramanick, Yale Song, Sayan Nag, Kevin~Qinghong Lin, Hardik Shah, Mike~Zheng Shou, Rama Chellappa, and Pengchuan Zhang.
\newblock {EgoVLPv2}: Egocentric video-language pre-training with fusion in the backbone.
\newblock In \emph{Proceedings of the IEEE/CVF International Conference on Computer Vision}, pages 5285--5297, 2023.

\bibitem[Radford et~al.(2021)Radford, Kim, Hallacy, Ramesh, Goh, Agarwal, Sastry, Askell, Mishkin, Clark, Krueger, and Sutskever]{radford2021clip}
Alec Radford, Jong~Wook Kim, Chris Hallacy, Aditya Ramesh, Gabriel Goh, Sandhini Agarwal, Girish Sastry, Amanda Askell, Pamela Mishkin, Jack Clark, Gretchen Krueger, and Ilya Sutskever.
\newblock Learning transferable visual models from natural language supervision.
\newblock In \emph{Proceedings of the 38th International Conference on Machine Learning}, pages 8748--8763. PMLR, 2021.

\bibitem[Sun et~al.(2026)Sun, Wu, Zhu, Chen, Jiang, Ji, Zhu, Shi, Wu, Wang, and Yang]{sun2026dualroute}
Yuyang Sun, Yongliang Wu, Xingyu Zhu, Yuxia Chen, Zhenxiang Jiang, Yangguang Ji, Wenbo Zhu, Yanxi Shi, Jay Wu, Shuo Wang, and Xu Yang.
\newblock Dual-route top-$k$ retrieval with 1v1 {VLM} reranking for the {CoVR-R}, 2026.
\newblock arXiv preprint and CoVR-R challenge technical report; peer review not established.

\bibitem[Thawakar et~al.(2024)Thawakar, Naseer, Anwer, Khan, Felsberg, Shah, and Khan]{thawakar2024ecde}
Omkar Thawakar, Muzammal Naseer, Rao~Muhammad Anwer, Salman Khan, Michael Felsberg, Mubarak Shah, and Fahad~Shahbaz Khan.
\newblock Composed video retrieval via enriched context and discriminative embeddings.
\newblock In \emph{Proceedings of the IEEE/CVF Conference on Computer Vision and Pattern Recognition}, pages 26896--26906, 2024.

\bibitem[Thawakar et~al.(2025)Thawakar, Demidov, Thawkar, Anwer, Shah, Khan, and Khan]{thawakar2025dense}
Omkar Thawakar, Dmitry Demidov, Ritesh Thawkar, Rao~Muhammad Anwer, Mubarak Shah, Fahad~Shahbaz Khan, and Salman Khan.
\newblock Beyond simple edits: Composed video retrieval with dense modifications.
\newblock In \emph{Proceedings of the IEEE/CVF International Conference on Computer Vision}, pages 20435--20444, 2025.

\bibitem[Thawakar et~al.(2026)Thawakar, Demidov, Potlapalli, Bogireddy, Gajjala, Lasheen, Anwer, and Khan]{thawakar2026covrr}
Omkar Thawakar, Dmitry Demidov, Vaishnav Potlapalli, Sai Prasanna Teja~Reddy Bogireddy, Viswanatha~Reddy Gajjala, Alaa~Mostafa Lasheen, Rao~Muhammad Anwer, and Fahad Khan.
\newblock {CoVR-R}: Reason-aware composed video retrieval.
\newblock In \emph{British Machine Vision Conference}, 2026.
\newblock Accepted; proceedings forthcoming.

\bibitem[Tu et~al.(2025)Tu, Jin, Liao, Luo, Wang, Shen, and Tao]{tu2025mllmguided}
Rong-Cheng Tu, Zhao Jin, Jingyi Liao, Xiao Luo, Yingjie Wang, Li Shen, and Dacheng Tao.
\newblock {MLLM}-guided {VLM} fine-tuning with joint inference for zero-shot composed image retrieval, 2025.
\newblock arXiv preprint arXiv:2505.19707.

\bibitem[Ventura et~al.(2024{\natexlab{a}})Ventura, Yang, Schmid, and Varol]{ventura2024covr}
Lucas Ventura, Antoine Yang, Cordelia Schmid, and G{\"u}l Varol.
\newblock {CoVR}: Learning composed video retrieval from web video captions.
\newblock In \emph{Proceedings of the AAAI Conference on Artificial Intelligence}, pages 5270--5279, 2024{\natexlab{a}}.

\bibitem[Ventura et~al.(2024{\natexlab{b}})Ventura, Yang, Schmid, and Varol]{ventura2024covr2}
Lucas Ventura, Antoine Yang, Cordelia Schmid, and G{\"u}l Varol.
\newblock {CoVR-2}: Automatic data construction for composed video retrieval.
\newblock \emph{IEEE Transactions on Pattern Analysis and Machine Intelligence}, 46\penalty0 (12):\penalty0 11409--11421, 2024{\natexlab{b}}.

\bibitem[Vo et~al.(2019)Vo, Jiang, Sun, Murphy, Li, Fei-Fei, and Hays]{vo2019tirg}
Nam Vo, Lu Jiang, Chen Sun, Kevin Murphy, Li-Jia Li, Li Fei-Fei, and James Hays.
\newblock Composing text and image for image retrieval---an empirical odyssey.
\newblock In \emph{Proceedings of the IEEE/CVF Conference on Computer Vision and Pattern Recognition}, pages 6439--6448, 2019.

\bibitem[Wang et~al.(2026)Wang, Liu, Jiao, Wang, Li, Li, Chen, and Liu]{wang2026visionbyprompt}
Hao Wang, Fang Liu, Licheng Jiao, Jiahao Wang, Shuo Li, Lingling Li, Puhua Chen, and Xu Liu.
\newblock Vision-by-prompt: Context-aware dual prompts for composed video retrieval.
\newblock \emph{Pattern Recognition}, 172:\penalty0 112378, 2026.

\bibitem[Wen et~al.(2026)Wen, Song, Zhang, Guan, Zhao, and Nie]{wen2026unicvr}
Haokun Wen, Xuemeng Song, Haoyu Zhang, Weili Guan, Xiangyu Zhao, and Liqiang Nie.
\newblock {UniCVR}: From alignment to reranking for unified zero-shot composed visual retrieval, 2026.
\newblock arXiv preprint arXiv:2604.20318.

\bibitem[Wu et~al.(2025)Wu, Qi, Wu, Sun, Wang, and Wang]{wu2025finecvr}
Yue Wu, Zhaobo Qi, Yiling Wu, Junshu Sun, Yaowei Wang, and Shuhui Wang.
\newblock Learning fine-grained representations through textual token disentanglement in composed video retrieval.
\newblock In \emph{International Conference on Learning Representations}, 2025.

\bibitem[Xu et~al.(2026)Xu, Zhang, Wu, Lu, Maneriker, Du, Li, and Liu]{xu2026videocritic}
Chenwei Xu, Jianshu Zhang, Shang Wu, Lie Lu, Pranav Maneriker, Fan Du, Manling Li, and Han Liu.
\newblock {VideoCritic}: Diagnosing and localizing reasoning errors in video-language models.
\newblock In \emph{CVPR Workshop on Video Large Language Models}, 2026.
\newblock Workshop publication; public author record and OpenReview PDF available.

\bibitem[Xu et~al.(2021)Xu, Ghosh, Huang, Okhonko, Aghajanyan, Metze, Zettlemoyer, and Feichtenhofer]{xu2021videoclip}
Hu Xu, Gargi Ghosh, Po-Yao Huang, Dmytro Okhonko, Armen Aghajanyan, Florian Metze, Luke Zettlemoyer, and Christoph Feichtenhofer.
\newblock {VideoCLIP}: Contrastive pre-training for zero-shot video-text understanding.
\newblock In \emph{Proceedings of the 2021 Conference on Empirical Methods in Natural Language Processing}, pages 6787--6800. Association for Computational Linguistics, 2021.

\bibitem[Zhang et~al.(2026{\natexlab{a}})Zhang, Zhang, Cao, Li, Li, Liu, and Wang]{zhang2026recovr}
Bingqing Zhang, Yi Zhang, Zhuo Cao, Yang Li, Xue Li, Jiajun Liu, and Sen Wang.
\newblock {ReCoVR}: Closing the loop in interactive composed video retrieval, 2026{\natexlab{a}}.
\newblock arXiv preprint arXiv:2605.09836.

\bibitem[Zhang et~al.(2026{\natexlab{b}})Zhang, Chen, Li, Fu, Wang, Nie, Wei, and Hu]{zhang2026relate}
Shiqi Zhang, Zhiwei Chen, Zixu Li, Zhiheng Fu, Wenbo Wang, Jiajia Nie, Yinwei Wei, and Yupeng Hu.
\newblock {RELATE}: Enhance composed video retrieval via minimal-redundancy hierarchical collaboration.
\newblock In \emph{2026 IEEE International Conference on Acoustics, Speech and Signal Processing}, pages 12132--12136, 2026{\natexlab{b}}.

\bibitem[Zhang et~al.(2025)Zhang, Zhang, Xie, Li, Dai, Long, Xie, Zhang, Li, and Zhang]{zhang2025gme}
Xin Zhang, Yanzhao Zhang, Wen Xie, Mingxin Li, Ziqi Dai, Dingkun Long, Pengjun Xie, Meishan Zhang, Wenjie Li, and Min Zhang.
\newblock Bridging modalities: Improving universal multimodal retrieval by multimodal large language models.
\newblock In \emph{Proceedings of the IEEE/CVF Conference on Computer Vision and Pattern Recognition}, pages 9274--9285, 2025.

\bibitem[Zheng and Georgescu(2026)]{zheng2026xaligner}
Yuqian Zheng and Mariana-Iuliana Georgescu.
\newblock {X-Aligner}: Composed visual retrieval without the bells and whistles.
\newblock In \emph{Proceedings of the IEEE/CVF Conference on Computer Vision and Pattern Recognition Workshops}, pages 6073--6082, 2026.

\bibitem[Zhu et~al.(2024)Zhu, Lin, Ning, Yan, Cui, Wang, Pang, Jiang, Zhang, Li, Zhang, Li, Liu, and Yuan]{zhu2024languagebind}
Bin Zhu, Bin Lin, Munan Ning, Yang Yan, Jiaxi Cui, Hongfa Wang, Yatian Pang, Wenhao Jiang, Junwu Zhang, Zongwei Li, Wancai Zhang, Zhifeng Li, Wei Liu, and Li Yuan.
\newblock {LanguageBind}: Extending video-language pretraining to n-modality by language-based semantic alignment.
\newblock In \emph{International Conference on Learning Representations}, 2024.

\end{thebibliography}
}

\clearpage
\maketitlesupplementary
\appendix
\setcounter{figure}{0}
\setcounter{table}{0}
\setcounter{equation}{0}
\renewcommand{\thefigure}{S\arabic{figure}}
\renewcommand{\thetable}{S\arabic{table}}
\renewcommand{\theequation}{S\arabic{equation}}
\section{Additional Method Details}
\label{app:method-details}

\subsection{Pipeline Overview}
\label{app:pipeline-overview}

This section gives the executable order, equality cases, and fallback invariants behind the conceptual cascade in the main paper. Let \(e_1\ge e_2\) be the two leading cosine similarities after source masking and \(r_1\ge r_2\) the leading candidate-relevance scores. All comparisons use unrounded values.

\textbf{Reusable gallery construction.} After applying any positive query limit, the evaluator reconstructs the target-only gallery from the retained examples. It validates the media, removes duplicate targets by stable identifier, resolves query-independent target selections, and embeds each distinct target exactly once. Each normalized video-only vector is stored in a gallery matrix. The benchmark reconstructs this matrix within an invocation; a deployment may persist the same target selections and vectors. Candidate-frame preparation is also query-independent and reusable, although every relevance call between a query and candidate remains online.

\textbf{Online execution.} Frame indices are resolved before model inference. A terse edit can activate target-description generation while the composed query and gallery are embedded. After source self-masking, the initial five gallery items form the candidate seed. A sufficiently separated embedding leader returns the coarse ranking directly; otherwise, provisional relevance scoring controls candidate expansion and the deferred description route. At most one generated description is reused across all candidates, and unchanged pair scores need not be recomputed in the final reranking pass. The scored head replaces the corresponding portion of the embedding order while the untouched tail remains complete. Finally, only a leading near-tie cluster can reach strict verification.

Skipping, failure, or unresolved output at any conditional stage retains the preceding valid order. Independent items and queries are processed concurrently. Within a query, expansion pages remain sequential because each leading score determines whether another page is needed; verifier calls are likewise sequential because the first accepted candidate terminates the scan.

\subsection{Stage-Aware Frame Allocation}
\label{app:frame-allocation}

Each downstream video input receives at most \(B=15\) selected frames; short clips and failed decodes can yield fewer, and verification contains separate source and candidate inputs. Global embedding uses equidistant samples. Reranking, decomposition, and verification instead use duration-adaptive novelty allocation computed from query-independent, low-resolution frame probes. The probe count may exceed the downstream budget and is accounted for separately from the selected frames consumed by a later-stage model.

For duration \(T\), short \([0,10)\), medium \([10,30)\), and long \([30,\infty)\) clips are probed at 8, 4, and 2 frames per second:

\begin{equation}
\label{eq:supp-dynamic-allocation}
\begin{aligned}
\rho(T) &=
\begin{cases}
8, & 0\le T<10,\\
4, & 10\le T<30,\\
2, & T\ge 30,
\end{cases}\\
M_{\mathrm{req}}
  &= \max\!\left(1,\operatorname{round}(T\rho(T))\right),\\
M&=|\mathcal{P}|\le\min(F,M_{\mathrm{req}}),\\
\delta_i &=
\begin{cases}
    \max(0,1-\langle h_i,h_{i-1}\rangle), & 2\le i\le M,\\
\delta_2, & i=1,\ M>1,\\
0, & i=1,\ M=1,
\end{cases}\\
p_i&=
\begin{cases}
\delta_i/\sum_j\delta_j, & \sum_j\delta_j>0,\\
1/M, & \sum_j\delta_j=0,
\end{cases}\\
w_i&=\lambda p_i+(1-\lambda)/M,\qquad \lambda=1.
\end{aligned}
\end{equation}

In Equation~\ref{eq:supp-dynamic-allocation}, \(F\) is the original-frame count, \(\mathcal{P}\) is the set of \(M\) decoded probes, and \(h_i\) is the normalized embedding of probe \(i\). The first probe inherits the second probe's novelty when multiple probes exist. If \(M\le B\), all probes are retained. Otherwise, quantiles \((k+0.5)/B\), for \(k=0,\ldots,B-1\), are mapped through the inverse CDF of \(w\); duplicate selections are removed, the highest-density unused probes backfill the budget, and the result is sorted in time. Zero novelty triggers the uniform fallback. This distributes frames over novelty mass rather than selecting only the \(B\) largest changes.

The benchmark presamples indices over the active source and gallery set. Configurations sharing the same clip and selection are deduplicated, target selections can be cached with the gallery, and a new source selection is online preparation. Every downstream request uses the selected original-frame indices with server-side resampling disabled. Final decoding, resizing, timestamp rendering where applicable, and JPEG packing remain lazy. The spatial settings are aspect-preserving long-side caps: embedding uses uniform 512-pixel frames, novelty probing uses 256-pixel frames, reranking and decomposition use selected 256-pixel frames, and verification uses selected 512-pixel frames.

\begin{table*}[!t]
\caption{Evaluation datasets and protocol. Every row uses the complete named public split, a target-video gallery, and source self-masking when the query source appears as a non-target item.}
\label{tab:dataset-protocol}
\centering
\begin{tabular}{@{}llll@{}}
\toprule
Dataset & Evaluation split & Gallery & Source self-mask \\
\midrule
WebVid-CoVR & Complete public test set & Target videos & Yes \\
Dense-WebVid-CoVR & Complete public test set & Target videos & Yes \\
CoVR-R & Official public split & Target videos & Yes \\
\bottomrule
\end{tabular}

\end{table*}

\subsection{Composed Query Embedding and Coarse Retrieval}
\label{app:embedding}

The composed-query encoder receives the source video and unchanged raw edit under a target-oriented retrieval instruction. The gallery encoder receives only a target video under a separate, explicitly empty instruction; the empty role is transmitted rather than omitted. Both outputs are L2-normalized, so their dot product is cosine similarity. The source side uses up to 15 equidistant frames at a 512-pixel long-side cap, and each target is encoded once under the same visual policy. If the source video occurs in the gallery as a non-target item, it is masked before sorting. Strict score ties therefore retain the minimum rank convention defined in \Cref{app:protocol}.

The five highest-scoring valid items seed later stages, while the complete order is retained. The compact embedding model is deliberately assigned the gallery-wide role: it must summarize the broad intent expressed by the source and edit and preserve candidate recall rather than make the final fine-grained decision. Its one-vector target representation can be stored and searched with exact or approximate nearest-neighbor machinery, allowing more specialized models to operate only on a routed candidate subset. The exact role instructions and model identity are reported in \Cref{app:interfaces}.

\subsection{Confidence-Gated Adaptive Reranking}\label{app:reranking}

The candidate scorer receives the raw edit or one decomposed target description together with a candidate video; the source video is absent. The selected reranker exposes a continuous relevance score for its hosted binary-relevance task. The client consumes that score directly and does not reconstruct token probabilities.

Reranking is bypassed only when \(e_1-e_2>0.25\); equality at \(0.25\) enters the stage. Otherwise, the initial five candidates are scored. While the best relevance score is strictly below \(0.70\), the next five items in embedding order are appended and scored. Equality at \(0.70\) stops expansion. At most nine additional pages are admitted, giving a maximum scored pool of \(5+9\times5=50\) items, or the gallery size if smaller.

Scored candidates are sorted in decreasing score with stable tie handling. The unscored tail retains its embedding-order position, and at most the reranked top ten are exposed to verification. A provisional pass supplies the confidence signal for deferred decomposition; the final pass reuses every candidate score whose text and video inputs are unchanged. Pages are sequential within a query, while different queries can be processed concurrently. A fired embedding-gap gate, missing scores, or query-level reranking failure preserves the complete embedding order and cannot activate downstream score-dependent routing.

\subsection{Conditional Query Decomposition}\label{app:decomposition}

The decomposer receives the source video and raw edit and requests one or two target-oriented sentences. Its candidate-independent output is generated at most once for each unique pairing of source and edit and reused across candidates and expansion rounds. In the final configuration it replaces only the reranker's text; embedding and verification continue to use the raw modification.

The early route counts ASCII letter, digit, or apostrophe sequences as words and activates when the count is strictly below 10; an edit of exactly 10 words remains raw. If no usable early description exists, the late route activates only after a valid provisional reranking pass with \(r_1\ge0.25\) and \((r_1-r_2)/r_1<0.50\). The score floor is inclusive and the relative-margin test is strict. When a second score is absent it is treated as zero; a missing or nonpositive leading score cannot satisfy the gate. The provisional pass includes any adaptive expansion, so its final observed leaders define the decision. An embedding-gap bypass or failed reranking provides no late-route evidence, while an unsuccessful early generation may be reconsidered by the late rule.

Frame selection completes before early generation begins. Early requests can run in a background worker while query and then gallery embeddings are computed, with parallelism within each block; generation joins before reranking. Empty or failed output falls back to the raw edit. The parser, response budget, cache boundary, and failed-item recovery policy appear in \Cref{app:interfaces} and \Cref{app:reproducibility}.

\begin{table*}[!t]
\caption{Task-trained or adapted approaches. These methods use CoVR-specific training, learned aggregation, or adaptation and therefore provide a contextual rather than resource-equivalent comparison; CA denotes learned cross-attention fusion. Metric cells retain the precision printed by each source; Avg. is computed from the displayed cutoffs and rounded to two decimals, with missing cutoffs not imputed. Within each dataset, \bestscore{boldface} and \underline{underlining} mark the best and second-best available values. \unverifiedtextmark{} marks WebVid-CoVR, which includes unverified machine-generated modification queries, often short or too generic~\cite{ventura2024covr2,thawakar2025dense,thawakar2026covrr}. \unreviewedmark{} denotes a method without independently verified peer-reviewed acceptance at the time of verification. ``Dense-CoVR'' in MoRe refers to BSE-CoVR; we retain the original method name.}
\label{tab:task-trained-results}
\centering
\begin{tabular}{@{}llrrrrr@{}}
\toprule
Dataset & Method & R@1 & R@5 & R@10 & R@50 & Avg. \\
\specialrule{\heavyrulewidth}{2pt}{2pt}
CoVR-R
  & BSE-CoVR~\cite{thawakar2025dense,thawakar2026covrr} & \bestscore{37.90} & \bestscore{57.67} & \bestscore{64.48} & \bestscore{79.47} & \bestscore{59.88} \\
\specialrule{\heavyrulewidth}{3pt}{2pt}
\multirow[t]{4}{*}{Dense-WebVid-CoVR}
  & CoVR-BLIP (CA)~\cite{ventura2024covr,thawakar2026covrr} & 35.60 & 60.80 & 70.31 & 87.05 & 63.44 \\
  & ECDE (CA)~\cite{thawakar2024ecde,thawakar2026covrr} & 39.20 & 64.40 & 75.56 & 88.90 & 67.02 \\
  & BSE-CoVR (CA)~\cite{thawakar2025dense,thawakar2026covrr,huang2026more} & \underline{48.08} & \underline{73.36} & \underline{81.06} & \underline{93.78} & \underline{74.07} \\
  & BSE-CoVR (Dense-trained)~\cite{thawakar2025dense} & \bestscore{71.26} & \bestscore{89.12} & \bestscore{94.56} & \bestscore{98.88} & \bestscore{88.46} \\
\specialrule{\heavyrulewidth}{3pt}{2pt}
\multirow[t]{7}{*}{WebVid-CoVR\unverifiedtextmark}
  & FDCA-BLIP (FineCVR pretrained)~\cite{wu2025finecvr} & 52.23 & 79.42 & 86.66 & 96.91 & 78.81 \\
  & CoVR-BLIP (WebVid adapted)~\cite{ventura2024covr,thawakar2024ecde} & 53.13 & 79.93 & 86.85 & 97.69 & 79.40 \\
  & FDCA-BLIP (WebVid adapted)~\cite{wu2025finecvr} & 54.80 & 82.27 & 89.84 & 97.70 & 81.15 \\
  & CoVR-BLIP-2~\cite{ventura2024covr2} & 59.82 & 83.84 & 91.28 & 98.24 & 83.30 \\
  & ECDE (WebVid adapted)~\cite{thawakar2024ecde} & 60.12 & 84.32 & 91.27 & \underline{98.72} & 83.61 \\
  & UNITE$_{\mathrm{instruct}}$ 7B\unreviewedmark~\cite{kong2025unite} & \bestscore{72.50} & \bestscore{90.80} & \bestscore{95.30} & \bestscore{99.50} & \bestscore{89.53} \\
  & UniCVR Stage II\unreviewedmark~\cite{wen2026unicvr} & \underline{66.77} & \underline{86.31} & \underline{91.88} & 98.29 & \underline{85.81} \\
\bottomrule
\end{tabular}
\end{table*}

\subsection{Confidence-Gated Candidate Verification}\label{app:verification}

The verifier receives the source video, raw edit, and one candidate video; a decomposed description is not reused. Its strict prompt requires the candidate to satisfy every modified condition relative to the source. Missing, contradicted, or uncertain conditions yield a negative verdict, whereas unconstrained attributes need not match.

Only the reranked top ten are considered. Verification requires \(r_1\ge0.15\) and a contiguous leading cluster of at least two candidates satisfying \((r_1-r_i)/r_1<0.25\). The score floor is inclusive, while equality at the relative gap excludes a candidate and terminates the prefix. Missing or nonpositive leading scores, a one-item cluster, or a score below the floor bypass the stage.

Eligible candidates are scanned in rank order. The first strict positive verdict is stably promoted to the leading position and terminates the scan; all other relative positions are preserved. An all-negative scan, an unparseable response, exhaustion of retries, or a bypass leaves the reranked order unchanged. The generative request, parser, reasoning budget, sampling defaults, and recovery policy are specified in \Cref{app:interfaces}.

\subsection{Timestamp-Aware Conditioning}\label{app:timestamp-conditioning}

Only decomposition and verification receive visible timestamps. After selection, decoding, and resizing, the client computes elapsed seconds as the original frame index divided by the original frame rate. It renders the value to two decimal places in the top-left corner using white glyphs with a black outline, before JPEG encoding. Embedding, reranking, and novelty probes receive no overlay.

Both generative prompts append the note: \textquotedblleft A timestamp is drawn in each frame for reference only and is not part of the video. Videos do not have to be of the same length.\textquotedblright{} The overlay and prompt note are treated as one timestamp-aware package. Accordingly, the measured ablation is attributed to the package rather than to either rendering or prompt wording in isolation.

\section{Additional Experiments and Analysis}\label{app:experiments}

\subsection{Experimental Setup}\label{app:experimental-setup}

\subsubsection{Datasets and Evaluation Protocol}\label{app:protocol}

WebVid-CoVR contains short automatically derived edits, Dense-WebVid-CoVR supplies denser attribute, action, and temporal descriptions, and CoVR-R combines WebVid with Something-Something-V2 reasoning examples. Main comparisons use each complete named public evaluation. Controlled studies retain their source identity: the Dense-WebVid-CoVR subset remains a dense-description study, whereas the CoVR-R controlled cohort (used for ablations and analysis) is identified simply as its WebVid subset and contains no Something-Something-V2 examples. Table~\ref{tab:dataset-protocol} summarizes the datasets and shared protocol.

Our evaluations use a target-video gallery and mask the query source when it reappears as a non-target item. Prior values in the comparison tables follow each cited source's reported benchmark protocol, which is not always explicit about source self-masking. R@\(K\) is the fraction of targets whose score rank is at most \(K\); exact score ties receive the optimistic minimum rank. MeanR3 averages R@1, R@5, and R@10, while the Avg. column in Table~\ref{tab:extended-results} also includes R@50. MRR is reported on the unit interval, and stage-prefixed metrics are diagnostic snapshots rather than substitutes for the final ranking. Our averages use unrounded values; prior averages are source-reported or derived from their published recalls.

\begin{table*}[!t]
\caption{Cumulative full-evaluation performance after composed embedding, adaptive reranking with conditional decomposition, and selective verification. Deltas compare adjacent stages within the same final run and are computed before display rounding.}
\label{tab:stage-progression}
\centering
\begin{tabular}{@{}llrrrrrr@{}}
\toprule
Dataset & Cumulative stage & R@1 & R@5 & R@10 & R@50 & Avg. & $\Delta$R@1 \\
\midrule
WebVid-CoVR & Embedding & 30.63 & 57.24 & 67.53 & 86.38 & 60.45 & \textemdash{} \\
 & + Rerank /\allowbreak{} decompose & 53.25 & 73.12 & 79.26 & 86.38 & 73.00 & +22.61 \\
 & + Verify & 54.58 & 73.12 & 79.26 & 86.38 & 73.34 & +1.33 \\
\midrule
Dense-WebVid-CoVR & Embedding & 59.65 & 84.19 & 90.49 & 97.26 & 82.90 & \textemdash{} \\
 & + Rerank /\allowbreak{} decompose & 88.57 & 95.15 & 96.40 & 97.26 & 94.34 & +28.92 \\
 & + Verify & 89.55 & 95.15 & 96.40 & 97.26 & 94.59 & +0.98 \\
\midrule
CoVR-R & Embedding & 56.64 & 79.08 & 84.55 & 94.95 & 78.81 & \textemdash{} \\
 & + Rerank /\allowbreak{} decompose & 93.32 & 94.27 & 94.61 & 94.95 & 94.29 & +36.67 \\
 & + Verify & 93.43 & 94.27 & 94.61 & 94.95 & 94.31 & +0.11 \\
\bottomrule
\end{tabular}

\end{table*}

\subsubsection{Implementation Details}\label{app:implementation-details}

No benchmark example updates model parameters. Qwen3-VL-Embedding-2B jointly encodes source video and edit while keeping target encodings video-only. Qwen3-VL-Reranker-8B scores a small initial pool and grows it in fixed increments while confidence remains low; a large embedding margin bypasses reranking. Qwen3.5-9B supplies conditional target-description generation and strict verification. Description generation is routed by edit concision or reranker ambiguity, while verification scans only a leading relative-score cluster and accepts the first strict match.

Embedding uses a uniform reusable view. Reranking and decomposition use novelty-weighted 256-pixel frames; verification uses separate source and candidate views at 512 pixels. A query-independent 256-pixel probe selects later-stage frames at duration-dependent rates, and decomposition and verification receive the timestamp overlay together with its explanatory prompt note. Tables~\ref{tab:full-configuration} and~\ref{tab:prompt-interfaces} in \Cref{app:interfaces} consolidate model identities, exact thresholds, frame limits, generation settings, parsers, retries, and reuse behavior.

Training-free methods form the primary comparison in Table~\ref{tab:main-results}; task-trained or adapted systems are separated in Table~\ref{tab:task-trained-results} because they are contextual rather than resource-equivalent. Prior results are source-reported on the named benchmarks, whereas our rows use the target-gallery and source-self-mask protocol described above. App.~\ref{app:routing} profiles the selected configuration on size-matched 1,000-query, 1,000-target cohorts with fixed profiling seeds and reports both serving-dependent time and model-call workload (not a cross-system latency comparison).

\subsection{Quantitative Results}\label{app:quantitative-results}

\subsubsection{Cross-Dataset Performance}\label{app:cross-dataset}

Table~\ref{tab:extended-results} shows that the 89.55/95.15/96.40/97.26 Dense-WebVid-CoVR recall profile yields 94.59 Avg.; the largest improvement is at the head of the ranking, while R@50 is already close to saturation. On CoVR-R, the 93.43/94.27/94.61/94.95 profile exceeds the other directly compared training-free values at every cutoff, with the largest margin again at R@1. The result improves both early discrimination and coverage without repeating gallery-wide reasoning.

WebVid-CoVR's modification candidates were machine-generated and its test examples were manually selected or filtered, but many retained edits remain short and underspecified~\cite{ventura2024covr2,thawakar2026covrr}. This annotation style may contribute to the remaining gap: literal similarity can exploit sparse surface cues, whereas the cascade can overinterpret an ambiguous request. Our 73.34 Avg. is less than one point below the best complete training-free average; at R@1, the source-reported MoRe row is higher under its reported benchmark protocol. The comparison reverses decisively on Dense-WebVid-CoVR, whose test modifications are fully manually verified and corrected~\cite{thawakar2025dense}: \methodname{} exceeds MoRe by 39.95, 28.15, and 18.50 points at R@1, R@5, and R@10, respectively. This higher-quality, more specific annotation regime better reflects detailed user intent, although aggregate results alone cannot identify why MoRe's relative standing reverses across the two benchmarks. Table~\ref{tab:task-trained-results} provides the separate task-trained or adapted comparison; on the denser benchmarks, the frozen cascade compares favorably with the displayed BSE-CoVR rows.

\subsubsection{Where the Gains Arise}\label{app:stage-gains}

Table~\ref{tab:stage-progression} shows a stable division of labor. Candidate reranking and routed decomposition produce the dominant R@1 gains, while verification adds 1.33, 0.98, and 0.11 points on WebVid-CoVR, Dense-WebVid-CoVR, and CoVR-R. Its smaller contribution is expected because it only resolves the leading ambiguity cluster. R@50 is unchanged after refinement because later stages reorder a bounded head and preserve the embedding tail; the reusable index therefore supplies coverage while specialists improve local ordering.

\begin{table*}[!t]
\caption{Extended embedding ablations on their stated datasets and subsets, covering query composition, weighted fusion, embedding capacity, and role-specific query/target instructions. Controlled rows are identified textually as the Dense-WebVid-CoVR subset or WebVid subset of CoVR-R.}
\label{tab:embedding-ablations}
\centering
\scriptsize
\setlength{\tabcolsep}{3.2pt}
\resizebox{\textwidth}{!}{%
\begin{tabular}{@{}llrrrrrrr@{}}
\toprule
Configuration & Dataset & R@1 & R@5 & R@10 & R@50 & M3 & M4 & MRR \\
\midrule
\multicolumn{9}{@{}l}{\emph{Query composition}} \\
Text only & Dense-WebVid-CoVR & 36.60 & 54.60 & 62.60 & 75.60 & 51.27 & 57.35 & 0.4547 \\
Source video only & Dense-WebVid-CoVR & 52.70 & 77.00 & 83.70 & 93.70 & 71.13 & 76.78 & 0.6327 \\
Weighted sum (text weight 0.25) & Dense-WebVid-CoVR & 57.60 & 77.40 & 83.50 & 93.20 & 72.83 & 77.92 & 0.6678 \\
Weighted sum (text weight 0.50) & Dense-WebVid-CoVR & 63.00 & 84.80 & 90.10 & 96.90 & 79.30 & 83.70 & 0.7275 \\
Weighted sum (text weight 0.75) & Dense-WebVid-CoVR & 57.90 & 80.80 & 87.20 & 96.20 & 75.30 & 80.53 & 0.6833 \\
Joint source-video + edit input & Dense-WebVid-CoVR & 67.20 & 85.80 & 89.20 & 96.00 & 80.73 & 84.55 & 0.7526 \\
\midrule
\multicolumn{9}{@{}l}{\emph{How does embedding-model capacity transfer across datasets?}} \\
Dense-WebVid-CoVR  /\allowbreak{}  8B & Dense-WebVid-CoVR & 67.20 & 85.80 & 89.20 & 96.00 & 80.73 & 84.55 & 0.7526 \\
Dense-WebVid-CoVR  /\allowbreak{}  2B & Dense-WebVid-CoVR & 69.60 & 90.10 & 94.20 & 98.80 & 84.63 & 88.17 & 0.7842 \\
WebVid-CoVR  /\allowbreak{}  8B & WebVid-CoVR & 44.20 & 68.60 & 76.50 & 89.00 & 63.10 & 69.58 & 0.5552 \\
WebVid-CoVR  /\allowbreak{}  2B & WebVid-CoVR & 47.70 & 75.00 & 83.70 & 95.90 & 68.80 & 75.58 & 0.5998 \\
CoVR-R WebVid-only subset  /\allowbreak{}  8B & CoVR-R WebVid-only & 73.80 & 89.80 & 93.60 & 98.20 & 85.73 & 88.85 & 0.8094 \\
CoVR-R WebVid-only subset  /\allowbreak{}  2B & CoVR-R WebVid-only & 72.50 & 92.30 & 95.70 & 98.80 & 86.83 & 89.83 & 0.8076 \\
\midrule
\multicolumn{9}{@{}l}{\emph{Do role-specific embedding instructions improve alignment?}} \\
Shared instruction & Dense-WebVid-CoVR & 69.60 & 90.10 & 94.20 & 98.80 & 84.63 & 88.17 & 0.7842 \\
Role-specific instructions & Dense-WebVid-CoVR & 73.10 & 91.50 & 95.30 & 98.90 & 86.63 & 89.70 & 0.8123 \\
\bottomrule
\end{tabular}

}
\end{table*}

\subsection{Qualitative Results}\label{app:qualitative}

\subsubsection{Candidate Evolution Through the Cascade}\label{app:candidate-evolution}

Figure~\ref{fig:first-page-teaser} in the main paper makes the stage roles visible without relying on trace identifiers. Coarse retrieval keeps several visually related water clips, reranking promotes both the target and a strong river distractor, and verification checks the remaining negative constraint. The wider scene is rejected because rocks and trees violate the request for no background objects; the close-up target satisfies the full edit. Figure~\ref{fig:qualitative-temporal-success} complements that progression with a temporal case: the candidate order remains unchanged after embedding and reranking, but the combined timestamp overlay and explanatory prompt let verification distinguish the complete train-window transformation from a passenger-only match. Together, the examples isolate broad recall, candidate ordering, and final constraint checking.

\subsubsection{Positive, Negative, and Ambiguous Retrievals}\label{app:retrieval-casebook}

\begin{figure}[tb]
\centering
\begin{tikzpicture}[
  qual image/.style={inner sep=.8pt,outer sep=0pt,line width=.7pt},
  qual pill/.style={draw=RetrievalQuery,fill=RetrievalQuery!7,
    rounded corners=1.5pt,text width=.78\columnwidth,align=center,
    inner xsep=2pt,inner ysep=2pt,font=\scriptsize},
  qual badge/.style={anchor=north west,text=white,font=\tiny\bfseries,
    inner xsep=2pt,inner ysep=1.2pt},
  qual reason/.style={anchor=north,font=\scriptsize\bfseries,align=center,
    text width=.43\columnwidth,inner sep=0pt},
  qual flow/.style={-{Latex[length=3pt,width=2.5pt]},line width=.7pt}
]
\node[qual image,draw=RetrievalNeutral] (source)
  {\includegraphics[width=.48\columnwidth]{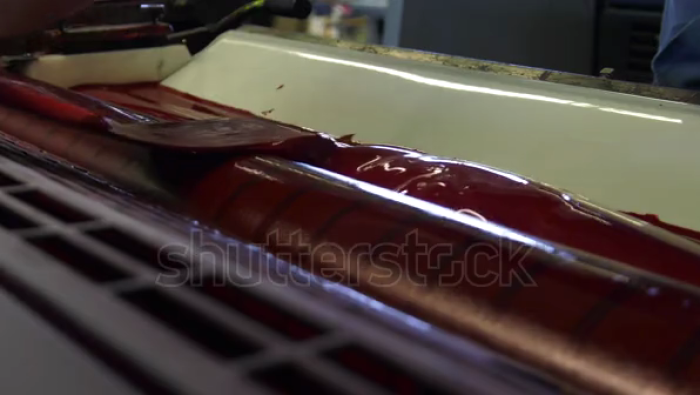}};
\node[qual pill,below=2.5pt of source] (edit)
  {\textbf{Edit:} change to blue};
\coordinate (fork) at ([yshift=-7pt]edit.south);
\node[qual image,draw=RetrievalPositive,densely dashed,anchor=north] (target)
  at ([xshift=-.235\columnwidth,yshift=-10pt]fork)
  {\includegraphics[width=.43\columnwidth]{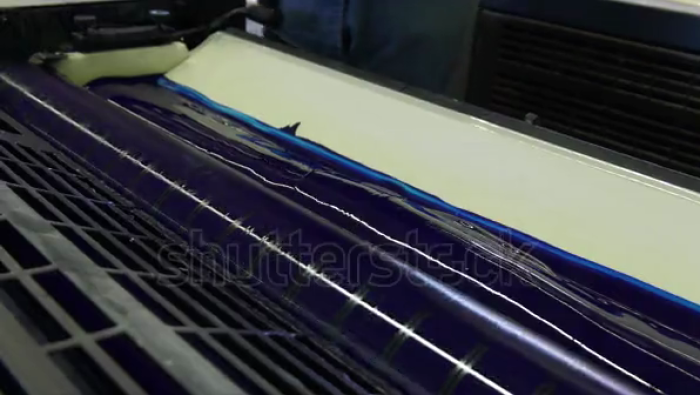}};
\node[qual image,draw=RetrievalNegative,anchor=north] (distractor)
  at ([xshift=.235\columnwidth,yshift=-10pt]fork)
  {\includegraphics[width=.43\columnwidth]{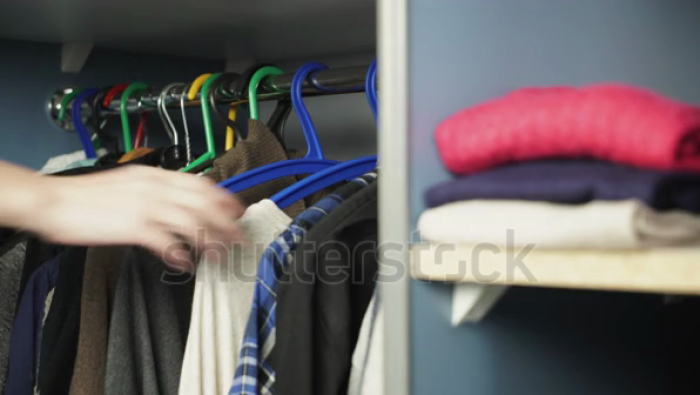}};

\draw[qual flow,RetrievalQuery] (source.south) -- (edit.north);
\draw[line width=.7pt,RetrievalNeutral] (edit.south) -- (fork);
\fill[RetrievalNeutral] (fork) circle[radius=.8pt];
\draw[qual flow,RetrievalPositive,densely dashed] (fork) -| (target.north);
\draw[qual flow,RetrievalNegative] (fork) -| (distractor.north);

\node[qual badge,fill=RetrievalNeutral]
  at ([xshift=.8pt,yshift=-.8pt]source.north west) {SOURCE};
\node[qual badge,fill=RetrievalPositive]
  at ([xshift=.8pt,yshift=-.8pt]target.north west)
  {$\checkmark$\;TARGET (MISSED)};
\node[qual badge,fill=RetrievalNegative]
  at ([xshift=.8pt,yshift=-.8pt]distractor.north west)
  {$\times$\;RETRIEVED DISTRACTOR};
\node[qual reason,text=RetrievalPositive]
  at ([yshift=-2pt]target.south) {Correct target; outside pool};
\node[qual reason,text=RetrievalNegative]
  at ([yshift=-2pt]distractor.south) {Blue content; wrong scene};
\end{tikzpicture}
\caption{\textbf{Sparse edits can fail before reasoning.} The correct factory process (green, dashed) remains outside the candidate pool, while the solid branch retrieves an unrelated blue clip. Later stages cannot recover an unseen target.}
\label{fig:qualitative-upstream-failure}
\end{figure}

An easy green-network example follows the shallow path because coarse retrieval already identifies the correct visual transformation, reranking preserves it, and verification is bypassed. At the other extreme, the sparse edit ``change to blue'' leaves the factory-process target outside the bounded pool, as shown in Figure~\ref{fig:qualitative-upstream-failure}; no later stage can score what coarse retrieval never admits. The corresponding dense edit specifies blue ink, a spatula, the factory setting, and background machinery, placing the same target inside the cascade and showing how query specificity controls upstream coverage.

\begin{table*}[!t]
\caption{Complete final target-gallery results with all recall cutoffs, MeanR3, MeanR4, MRR, MedianRank, and usable-media coverage.}
\label{tab:extended-results}
\centering
\begin{tabular}{@{}lrrrrrrrrr@{}}
\toprule
Dataset & R@1 & R@5 & R@10 & R@50 & M3 & M4 & MRR & MedR & Coverage \\
\midrule
WebVid-CoVR & 54.58 & 73.12 & 79.26 & 86.38 & 68.99 & 73.34 & 0.6278 & 1 & 100.00\% \\
Dense-WebVid-CoVR & 89.55 & 95.15 & 96.40 & 97.26 & 93.70 & 94.59 & 0.9211 & 1 & 100.00\% \\
CoVR-R & 93.43 & 94.27 & 94.61 & 94.95 & 94.10 & 94.31 & 0.9393 & 1 & 100.00\% \\
\bottomrule
\end{tabular}

\end{table*}

Some negative samples, like a campfire failure, expose a different limitation. The selected alternative contains orange-yellow flames, logs, embers, and a dark background and plausibly satisfies the request, yet the single-positive protocol marks only a near-duplicate as correct. This should be treated as annotation ambiguity rather than an unequivocal semantic error. Strict verification is corrective on balance but can still reject a correct near-tied candidate.

\subsubsection{Relation to the Main Qualitative Results}\label{app:main-qualitative-relation}

Candidate expansion rescues a bathtub transformation that lies outside the small initial seed: one confidence-triggered enlargement admits the target and reranking promotes it. This is a coverage correction rather than a semantic rewrite, because the original query remains unchanged. A controlled green-line diagnostic addresses the complementary bottleneck: decomposition enriches ``replace the white lines and dots with green'' with the missing moving-network context. That diagnostic supplies the generated text to both retrieval and reranking, whereas the final system uses the more stable reranker-only placement. The two cases therefore separate uncertainty about candidate depth from uncertainty about what the edit describes.

\subsection{Ablation Studies}\label{app:ablations}

\subsubsection{Embedding Stage}\label{app:embedding-ablations}

\paragraph{Query Strategy Analysis.}
Table~\ref{tab:embedding-ablations} shows that joint encoding of the source video and edit yields the best R@1 and MeanR3 in the complete fusion sweep, while equal-weight late fusion gives the best R@10. The joint encoder therefore favors the retrieval head without sacrificing a reusable video-only gallery.

\paragraph{Model Size.}
The compact embedder improves MeanR3 on the Dense-WebVid-CoVR, WebVid-CoVR, and WebVid subset of CoVR-R studies; only CoVR-R R@1 is slightly lower. More gallery-wide parameters are therefore not uniformly better.

\paragraph{Query/Target Decoupled Instructions.}
Role-specific instructions improve R@1 from 69.60 to 73.10 over a shared instruction. The query carries the transformation, while the target role remains a neutral video representation.

\subsubsection{Reranker Stage}\label{app:reranker-ablations}

\paragraph{Query Strategy Analysis.}
Table~\ref{tab:reranker-ablations} shows that adding text-conditioned candidate scoring raises Dense-WebVid-CoVR R@1 from 73.10 to 86.10. Supplying source video with the edit reaches 90.00, but the selected text interface reuses one raw or decomposed description across every candidate and expansion round.

\paragraph{Embedding-Based Reranker Gating.}
A large separation between the two leading embedding scores bypasses reranking for 7.50\% of queries without changing any reported accuracy metric, avoiding redundant candidate work on clear cases.

\paragraph{Reranker Pool Size.}
A small fixed seed is sufficient only when paired with conditional growth. It keeps confident cases shallow while allowing uncertain cases to reach deeper candidates, rather than asserting that one pool size is optimal for every query.

\paragraph{Score-Based Candidate Pool Expansion.}
Confidence-triggered growth improves R@1 from 86.10 to 89.50 and MeanR3 from 90.97 to 94.97. The matched combined-query study shows the same direction, confirming that the gain is a pool-depth effect rather than an artifact of the text interface.

\paragraph{Reranker Model Size.}
On the WebVid subset of CoVR-R, the larger candidate scorer raises R@1 from 96.57 to 98.17 while reducing both expansion and verification routing. Here, added reranker capacity improves local ordering and reduces downstream ambiguity. Compact decomposition frames preserve accuracy, while compact reranker frames trade a small R@1 decrease for lower repeated visual cost.

\subsubsection{Decomposer Stage}\label{app:decomposer-ablations}

\paragraph{Text Decomposition for Reranker.}
Table~\ref{tab:routing-tradeoffs} shows that target-oriented decomposition raises Dense-WebVid-CoVR R@1 from 89.50 to 94.30 and produces an even larger gain on WebVid-CoVR. Supplying the generated description to embedding has mixed cross-dataset effects, so the final interface confines it to reranking.

\paragraph{Decomposition Gating.}
Always-on generation gives the highest controlled accuracy, concise-edit routing is inexpensive but misses many ambiguous cases, and ambiguity routing captures most of the benefit. The selected union of concise and ambiguous edits activates for 26.50\% of queries and retains most of the always-on gain.

\subsubsection{Verifier Stage}\label{app:verifier-ablations}

\begin{table*}[!t]
\caption{Extended reranker ablations with full retrieval metrics and workload fields. Rows isolate text versus combined query input, 2B versus 8B capacity, the strict embedding-difference bypass, score-gated candidate expansion, and later-stage resolution using matched target-only protocols.}
\label{tab:reranker-ablations}
\centering
\scriptsize
\setlength{\tabcolsep}{3.2pt}
\resizebox{\textwidth}{!}{%
\begin{tabular}{@{}llrrrrrrrl@{}}
\toprule
Configuration & Dataset & R@1 & R@5 & R@10 & R@50 & M3 & M4 & MRR & Workload \\
\midrule
\multicolumn{10}{@{}l}{\emph{How much does candidate-conditioned reranking repair coarse retrieval?}} \\
Coarse embedding ranking & Dense-WebVid-CoVR & 73.10 & 91.50 & 95.30 & 98.90 & 86.63 & 89.70 & 0.8123 & \textemdash{} \\
+ 8B text reranker & Dense-WebVid-CoVR & 86.10 & 91.50 & 95.30 & 98.90 & 90.97 & 92.95 & 0.8933 & \textemdash{} \\
\midrule
\multicolumn{10}{@{}l}{\emph{Which query representation should the reranker score?}} \\
Edit text & Dense-WebVid-CoVR & 86.10 & 91.50 & 95.30 & 98.90 & 90.97 & 92.95 & 0.8933 & \textemdash{} \\
Source video + edit & Dense-WebVid-CoVR & 90.00 & 91.50 & 95.30 & 98.90 & 92.27 & 93.92 & 0.9149 & \textemdash{} \\
\midrule
\multicolumn{10}{@{}l}{\emph{Can high-margin candidates bypass redundant reranking?}} \\
Difference gate disabled & Dense-WebVid-CoVR & 86.10 & 91.50 & 95.30 & 98.90 & 90.97 & 92.95 & 0.8933 & \textemdash{} \\
Difference gate enabled & Dense-WebVid-CoVR & 86.10 & 91.50 & 95.30 & 98.90 & 90.97 & 92.95 & 0.8933 & skip 7.50\% \\
\midrule
\multicolumn{10}{@{}l}{\emph{When should adaptive reranking expand the candidate pool?}} \\
Text query  /\allowbreak{}  fixed pool & Dense-WebVid-CoVR & 86.10 & 91.50 & 95.30 & 98.90 & 90.97 & 92.95 & 0.8933 & expand 0.00\%, expanded 0, rescues 0, pages 0.00 \\
Text query  /\allowbreak{}  adaptive expansion & Dense-WebVid-CoVR & 89.50 & 97.20 & 98.20 & 98.90 & 94.97 & 95.95 & 0.9316 & expand 53.10\%, 57 rescues, mean depth 8.64 \\
Combined query  /\allowbreak{}  fixed pool & Dense-WebVid-CoVR & 90.00 & 91.50 & 95.30 & 98.90 & 92.27 & 93.92 & 0.9149 & expand 0.00\%, expanded 0, rescues 0, pages 0.00 \\
Combined query  /\allowbreak{}  adaptive expansion & Dense-WebVid-CoVR & 95.90 & 98.00 & 98.30 & 98.90 & 97.40 & 97.78 & 0.9696 & expand 17.40\%, 61 rescues, mean depth 6.85 \\
\midrule
\multicolumn{10}{@{}l}{\emph{How does reranker capacity affect accuracy and routed workload?}} \\
2B reranker & CoVR-R WebVid subset & 96.57 & 98.83 & 98.98 & 99.12 & 98.13 & 98.37 & 0.9765 & expand 82.47\%, verify 42.88\%, 861 calls \\
8B reranker & CoVR-R WebVid subset & 98.17 & 98.83 & 98.90 & 99.12 & 98.64 & 98.76 & 0.9849 & expand 11.40\%, verify 3.51\%, 61 calls \\
\midrule
\multicolumn{10}{@{}l}{\emph{Can later reasoning retain accuracy with compact frames?}} \\
Decomposition frames: 720 px & Dense-WebVid-CoVR & 94.30 & 98.40 & 98.60 & 98.90 & 97.10 & 97.55 & 0.9618 & \textemdash{} \\
Decomposition frames: 256 px & Dense-WebVid-CoVR & 94.30 & 98.50 & 98.80 & 98.90 & 97.20 & 97.62 & 0.9627 & \textemdash{} \\
\bottomrule
\end{tabular}

}
\end{table*}

\paragraph{Verifier Stage After Reranking.}
Ungated verification improves R@1 from 93.60 to 94.20, confirming that a final multimodal comparison can repair residual candidate-order errors.

\paragraph{Verifier Gating.}
The relative-score and score-floor gate reduces verdict work by more than 90\% with only a small accuracy change. Restricting the scan to a leading ambiguity cluster is therefore preferable to verifying every candidate set.

\paragraph{Verifier Strict Prompt.}
Under the matched gate, strict full-condition checking raises R@1 from 94.00 to 94.50. The result supports a strict, confidence-gated, top-down verifier under the fixed interface comprising the source, edit, and candidate.

\begin{table*}[!p]
\caption{Trade-offs between accuracy and workload for decomposition and verification on the Dense-WebVid-CoVR subset, comparing always-on, concise-edit, ambiguity-based, combined, lenient, and strict routing variants.}
\label{tab:routing-tradeoffs}
\centering
\scriptsize
\setlength{\tabcolsep}{3.2pt}
\resizebox{\textwidth}{!}{%
\begin{tabular}{@{}lrrrrrrrlrrr@{}}
\toprule
Configuration & R@1 & R@5 & R@10 & R@50 & M3 & M4 & MRR & Routed & Rescues & Calls & Saved \\
\midrule
\multicolumn{12}{@{}l}{\emph{Decomposition}} \\
Decomposition disabled & 89.50 & 97.20 & 98.20 & 98.90 & 94.97 & 95.95 & 0.9316 & 0.00\% & \textemdash{} & \textemdash{} & \textemdash{} \\
Decompose every query & 94.30 & 98.50 & 98.80 & 98.90 & 97.20 & 97.62 & 0.9627 & 100.00\% & \textemdash{} & \textemdash{} & \textemdash{} \\
Route concise edits & 91.00 & 98.00 & 98.50 & 98.90 & 95.83 & 96.60 & 0.9424 & 4.50\% & \textemdash{} & \textemdash{} & \textemdash{} \\
Route ambiguous rankings & 92.90 & 97.80 & 98.40 & 98.90 & 96.37 & 97.00 & 0.9526 & 24.20\% & 47 & \textemdash{} & \textemdash{} \\
Route concise or ambiguous queries & 93.60 & 98.20 & 98.60 & 98.90 & 96.80 & 97.32 & 0.9579 & 26.50\% & 39 & \textemdash{} & \textemdash{} \\
\midrule
\multicolumn{12}{@{}l}{\emph{Verification}} \\
Verification disabled & 93.60 & 98.20 & 98.60 & 98.90 & 96.80 & 97.32 & 0.9579 & 0.00\% & \textemdash{} & \textemdash{} & \textemdash{} \\
Verify every candidate set & 94.20 & 98.30 & 98.60 & 98.90 & 97.03 & 97.50 & 0.9616 & 100.00\% & \textemdash{} & 1124 & \textemdash{} \\
Confidence-gated verification & 94.00 & 98.20 & 98.60 & 98.90 & 96.93 & 97.42 & 0.9599 & 7.80\% & \textemdash{} & 84 & 92.53\% \\
Confidence-gated strict verification & 94.50 & 98.20 & 98.60 & 98.90 & 97.10 & 97.55 & 0.9625 & 7.80\% & \textemdash{} & 104 & 90.75\% \\
\bottomrule
\end{tabular}

}
\end{table*}

\subsubsection{Frame Sampling}\label{app:frame-sampling-ablations}

\paragraph{Dynamic Sampling.}
Table~\ref{tab:frame-policy} reports that all-stage novelty-weighted allocation improves the matched Dense-WebVid-CoVR result from 94.40 to 95.00 R@1 and improves every reported recall cutoff. Figure~\ref{fig:dynamic-sampling-showcase} illustrates how the policy reallocates views. With later stages already dynamic, changing embedding from uniform to dynamic accounts for the remaining 0.30-point R@1 gain; the selected balanced profile keeps gallery-wide embedding uniform and applies dynamic views only to candidate and reasoning stages.

\paragraph{Hyperparameters for Dynamic Frame Sampling.}
Short, medium, and long clips are probed at 8, 4, and 2 frames per second, respectively, and pure novelty weighting performs best in the measured sweep. Probe work remains separate from the 15-frame model-input ceiling. Table~\ref{tab:duration-conditioned} shows that the aggregate gain is not uniform: shorter groups benefit, whereas long clips expose heterogeneity and small-sample uncertainty in the long-duration tail.

\begin{table*}[!p]
\caption{Aggregate frame-policy ablations under a fixed 15-frame output budget per video input. The evaluated dynamic package improves the matched aggregate result at every reported recall cutoff; placement and novelty-mixing rows record the measured accuracy/compute choices. Probe work is accounted separately from the selected output.}
\label{tab:frame-policy}
\centering
\scriptsize
\setlength{\tabcolsep}{3.2pt}
\resizebox{\textwidth}{!}{%
\begin{tabular}{@{}llrrrrrrr@{}}
\toprule
Configuration & Dataset & R@1 & R@5 & R@10 & R@50 & M3 & M4 & MRR \\
\midrule
\multicolumn{9}{@{}l}{\emph{Does dynamic sampling improve aggregate retrieval?}} \\
Uniform sampling at every stage & Dense-WebVid-CoVR & 94.40 & 98.20 & 98.60 & 98.90 & 97.07 & 97.52 & 0.9619 \\
Dynamic sampling at every stage & Dense-WebVid-CoVR & 95.00 & 98.70 & 98.80 & 99.20 & 97.50 & 97.92 & 0.9663 \\
\midrule
\multicolumn{9}{@{}l}{\emph{Where should dynamic allocation enter the cascade?}} \\
Uniform embedding; dynamic reasoning stages & Dense-WebVid-CoVR & 94.70 & 98.40 & 98.50 & 98.90 & 97.20 & 97.62 & 0.9635 \\
Dynamic sampling at every stage & Dense-WebVid-CoVR & 95.00 & 98.70 & 98.80 & 99.20 & 97.50 & 97.92 & 0.9663 \\
\midrule
\multicolumn{9}{@{}l}{\emph{How should novelty and uniform density be mixed?}} \\
Novelty-mixing coefficient 0.8 & Dense-WebVid-CoVR & 93.40 & 98.00 & 98.20 & 98.90 & 96.53 & 97.12 & 0.9548 \\
Novelty-mixing coefficient 0.9 & Dense-WebVid-CoVR & 94.30 & 98.10 & 98.50 & 98.90 & 96.97 & 97.45 & 0.9611 \\
Novelty-mixing coefficient 1.0 & Dense-WebVid-CoVR & 94.70 & 98.40 & 98.50 & 98.90 & 97.20 & 97.62 & 0.9635 \\
\bottomrule
\end{tabular}

}
\end{table*}

\begin{table*}[!p]
\caption{Target-duration analysis for matched frame policies. Dynamic allocation helps the aggregate result and the shorter-duration groups, while the long-duration groups expose heterogeneity and greater small-sample uncertainty in the long-duration tail.}
\label{tab:duration-conditioned}
\centering
\scriptsize
\setlength{\tabcolsep}{3.2pt}
\resizebox{\textwidth}{!}{%
\begin{tabular}{@{}llrrrrrr@{}}
\toprule
Policy & Target duration & R@1 & R@5 & R@10 & R@50 & M3 & MRR \\
\midrule
Uniform at all stages & $[0,10)$ s & 94.50 & 97.25 & 97.71 & 98.17 & 96.48 & 0.9595 \\
Uniform at all stages & $[10,30)$ s & 94.55 & 99.00 & 99.43 & 99.57 & 97.66 & 0.9659 \\
Uniform at all stages & $[30,60)$ s & 92.21 & 93.51 & 93.51 & 94.81 & 93.07 & 0.9275 \\
Uniform at all stages & $[60,\infty)$ s & 100.00 & 100.00 & 100.00 & 100.00 & 100.00 & 1.0000 \\
\midrule
Uniform embedding; dynamic reasoning & $[0,10)$ s & 95.41 & 97.25 & 97.25 & 98.17 & 96.64 & 0.9627 \\
Uniform embedding; dynamic reasoning & $[10,30)$ s & 95.12 & 99.28 & 99.43 & 99.57 & 97.94 & 0.9701 \\
Uniform embedding; dynamic reasoning & $[30,60)$ s & 88.31 & 93.51 & 93.51 & 94.81 & 91.77 & 0.9030 \\
Uniform embedding; dynamic reasoning & $[60,\infty)$ s & 100.00 & 100.00 & 100.00 & 100.00 & 100.00 & 1.0000 \\
\midrule
Dynamic at all stages & $[0,10)$ s & 96.33 & 98.17 & 98.17 & 99.08 & 97.55 & 0.9716 \\
Dynamic at all stages & $[10,30)$ s & 95.41 & 99.43 & 99.57 & 99.71 & 98.13 & 0.9719 \\
Dynamic at all stages & $[30,60)$ s & 87.01 & 93.51 & 93.51 & 94.81 & 91.34 & 0.8967 \\
Dynamic at all stages & $[60,\infty)$ s & 100.00 & 100.00 & 100.00 & 100.00 & 100.00 & 1.0000 \\
\bottomrule
\end{tabular}

}
\end{table*}

\begin{figure*}[!p]
\centering
\setlength{\tabcolsep}{2pt}
\begin{tabular}{cccc}
\includegraphics[width=.24\textwidth]{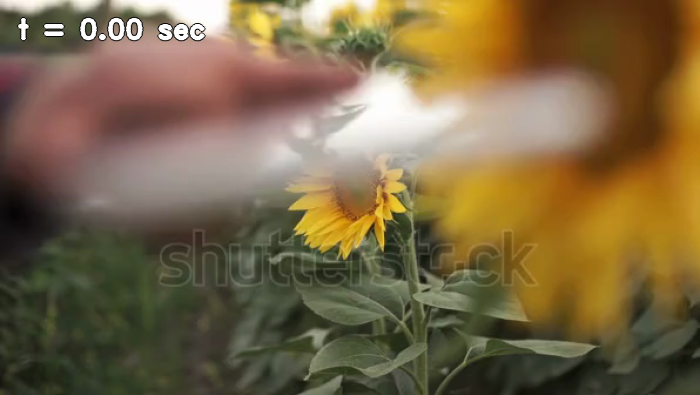} &
\includegraphics[width=.24\textwidth]{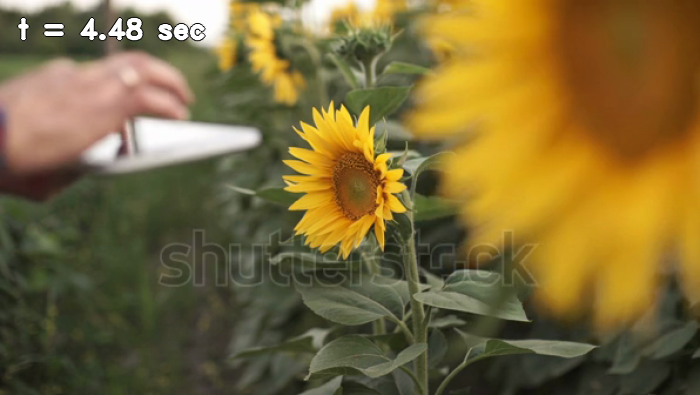} &
\includegraphics[width=.24\textwidth]{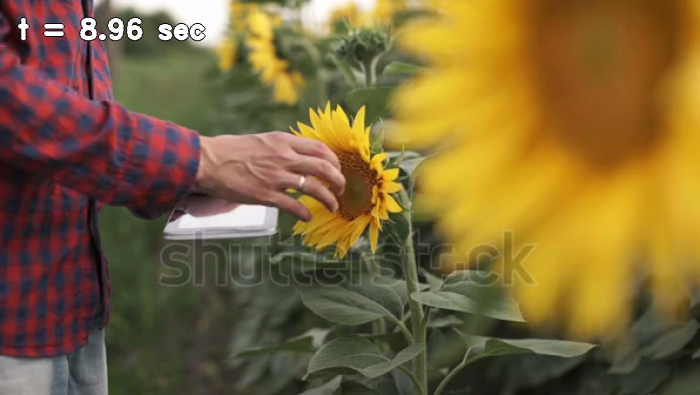} &
\includegraphics[width=.24\textwidth]{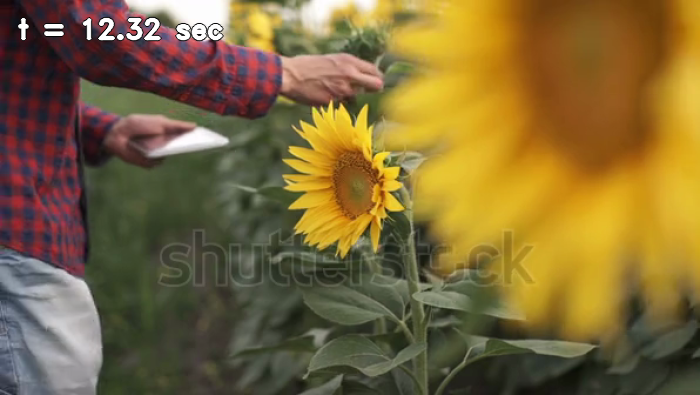}
\end{tabular}
\caption{\textbf{Timestamp overlay.} Selected frames carry elapsed seconds in the top-left corner, giving the reasoning stages an absolute temporal cue even when novelty-weighted samples are irregularly spaced. This visualization is separate from the dynamic-allocation comparison in Figure~\ref{fig:dynamic-sampling-showcase}.}
\label{fig:timestamp-overlay}
\end{figure*}

\paragraph{Frame Timestamp Overlay.}
Figure~\ref{fig:timestamp-overlay} visualizes the elapsed-time overlay, which is evaluated together with its explanatory prompt note. Table~\ref{tab:temporal-ablations} shows that enabling the package changes R@1 from 94.70 to 95.00 and MRR from 0.9652 to 0.9663, while R@50 is unchanged. The result supports the combined temporal cue and does not isolate overlay rendering from prompt wording.

\begin{table*}[!p]
\caption{Temporal ablations for the positive aggregate dynamic-sampling result and the combined timestamp-aware package. The timestamp rows jointly change decomposer overlays, verifier overlays, and the timestamp note; they are not component-wise causal estimates.}
\label{tab:temporal-ablations}
\centering
\scriptsize
\setlength{\tabcolsep}{3.2pt}
\resizebox{\textwidth}{!}{%
\begin{tabular}{@{}lrrrrrrrr@{}}
\toprule
Configuration & R@1 & R@5 & R@10 & R@50 & M3 & M4 & MRR & $\Delta$R@1 \\
\midrule
\multicolumn{9}{@{}l}{\emph{Does the dynamic frame-policy package improve aggregate retrieval?}} \\
Uniform frame allocation & 94.40 & 98.20 & 98.60 & 98.90 & 97.07 & 97.52 & 0.9619 & \textemdash{} \\
Dynamic frame-policy package & 95.00 & 98.70 & 98.80 & 99.20 & 97.50 & 97.92 & 0.9663 & +0.60 \\
\midrule
\multicolumn{9}{@{}l}{\emph{Does timestamp-aware reasoning improve temporal grounding?}} \\
Timestamp-aware package disabled & 94.70 & 98.80 & 98.90 & 99.20 & 97.47 & 97.90 & 0.9652 & \textemdash{} \\
Timestamp-aware package enabled & 95.00 & 98.70 & 98.80 & 99.20 & 97.50 & 97.92 & 0.9663 & +0.30 \\
\bottomrule
\end{tabular}

}
\end{table*}

Table~\ref{tab:extended-ablations} reports the cross-dataset frame-count and decomposition-placement extensions.

\begin{table*}[!p]
\caption{Cross-dataset extensions for frame count, decomposition activation, and decomposition placement. Full CoVR-R WebVid and Something-Something-V2 subset rows retain their actual subset identities; WebVid-CoVR and Dense-WebVid-CoVR rows remain separately labeled. The results show domain-dependent frame-count preferences and mixed embedding-side decomposition effects, supporting the conservative 15-frame shared ceiling and reranker-only final placement.}
\label{tab:extended-ablations}
\centering
\scriptsize
\setlength{\tabcolsep}{3.2pt}
\resizebox{\textwidth}{!}{%
\begin{tabular}{@{}llrrrrrrr@{}}
\toprule
Configuration & Dataset & R@1 & R@5 & R@10 & R@50 & M3 & M4 & MRR \\
\midrule
\multicolumn{9}{@{}l}{\emph{Frame budget}} \\
15 frames at all stages & CoVR-R WebVid subset & 97.15 & 97.81 & 97.88 & 98.17 & 97.61 & 97.75 & 0.9749 \\
7 frames at all stages & CoVR-R WebVid subset & 94.96 & 95.69 & 95.76 & 96.13 & 95.47 & 95.64 & 0.9534 \\
10 embedding frames; 15 later-stage frames & CoVR-R WebVid subset & 95.69 & 96.20 & 96.20 & 96.49 & 96.03 & 96.15 & 0.9597 \\
\midrule
\multicolumn{9}{@{}l}{\emph{Something-Something-V2 frame count}} \\
15 embedding frames & CoVR-R Something-Something-V2 subset & 78.10 & 78.67 & 78.67 & 79.63 & 78.48 & 78.77 & 0.7857 \\
10 embedding frames & CoVR-R Something-Something-V2 subset & 82.95 & 83.78 & 83.84 & 84.61 & 83.52 & 83.80 & 0.8352 \\
5 embedding frames & CoVR-R Something-Something-V2 subset & 86.65 & 87.48 & 87.48 & 87.93 & 87.21 & 87.39 & 0.8720 \\
\midrule
\multicolumn{9}{@{}l}{\emph{Decomposition activation}} \\
Decomposition off & WebVid-CoVR & 34.78 & 55.63 & 65.02 & 82.75 & 51.81 & 59.55 & 0.4448 \\
Reranker-side decomposition & WebVid-CoVR & 55.63 & 73.47 & 78.25 & 82.75 & 69.12 & 72.53 & 0.6357 \\
Decomposition off & Dense-WebVid-CoVR & 86.58 & 92.76 & 93.46 & 94.36 & 90.93 & 91.79 & 0.8935 \\
Reranker-side decomposition & Dense-WebVid-CoVR & 88.65 & 93.35 & 93.82 & 94.36 & 91.94 & 92.54 & 0.9084 \\
Decomposition off & CoVR-R WebVid subset & 96.49 & 97.95 & 98.10 & 98.17 & 97.52 & 97.68 & 0.9720 \\
Reranker-side decomposition & CoVR-R WebVid subset & 97.15 & 97.81 & 97.88 & 98.17 & 97.61 & 97.75 & 0.9749 \\
\midrule
\multicolumn{9}{@{}l}{\emph{Decomposition placement}} \\
Reranker only & Dense-WebVid-CoVR & 88.65 & 93.35 & 93.82 & 94.36 & 91.94 & 92.54 & 0.9084 \\
Embedding + reranker & Dense-WebVid-CoVR & 88.14 & 92.95 & 93.27 & 93.93 & 91.45 & 92.07 & 0.9038 \\
Reranker only & WebVid-CoVR & 55.63 & 73.47 & 78.25 & 82.75 & 69.12 & 72.53 & 0.6357 \\
Embedding + reranker & WebVid-CoVR & 57.55 & 76.76 & 83.06 & 89.16 & 72.46 & 76.63 & 0.6629 \\
\bottomrule
\end{tabular}

}
\end{table*}


\subsection{Efficiency and Additional Analysis}\label{app:routing}

\begin{table*}[!t]
\caption{Controlled inference cost for the selected configuration on size-matched 1,000-query, 1,000-target cohorts. Online busy time sums decomposition, query embedding, reranking, and verification model intervals; online wall is their interval union under 16-worker execution. Evaluator wall spans the evaluation routine after setup and before reporting and persistence, additionally including presampling, gallery preparation, retrieval, and analysis. Cross-run caches are disabled and in-run reuse is retained. Calls count issued model inferences, with gated skips counted as zero. CoVR-R denotes its controlled WebVid-subset cohort.}
\label{tab:efficiency}
\centering
\scriptsize
\textbf{(a) Measured execution and online compute}\\[-1pt]
\begin{tabular}{@{}lrrrrrr@{}}
\toprule
Dataset & \multicolumn{3}{c}{Time (s)} & \multicolumn{2}{c}{Online calls} \\
\cmidrule(lr){2-4}\cmidrule(l){5-6}
& Mean busy/query & Online wall & Evaluator wall & Mean/query & Total \\
\midrule
WebVid-CoVR       & 69.751 & 4416.26 & 5330.73 & 76.264 & 76{,}264 \\
Dense-WebVid-CoVR & 38.630 & 2488.34 & 3233.29 & 50.514 & 50{,}514 \\
CoVR-R            & 17.842 & 1204.25 & 2012.57 & 23.227 & 23{,}227 \\
\bottomrule
\end{tabular}

\vspace{4pt}
\textbf{(b) Routing activity on size-matched cohorts}\\[-1pt]
\begin{tabular}{@{}lrrrr@{}}
\toprule
Dataset & Embedding-gap bypass & Expanded & Decomposed & Verified \\
\midrule
WebVid-CoVR       & 7.3\% & 75.4\% & 99.6\% & 16.1\% \\
Dense-WebVid-CoVR & 7.6\% & 44.2\% & 26.1\% &  6.5\% \\
CoVR-R            & 7.7\% & 17.4\% & 11.1\% &  3.6\% \\
\bottomrule
\end{tabular}

\end{table*}

\begin{table*}[!t]
\caption{Stage and reusable-preparation breakdown for the controlled profiles in Table~\ref{tab:efficiency}. Stage cells report the mean busy seconds and issued model calls per query, including zero cost when a conditional stage is bypassed. Temporal selection is query-independent and reports one low-resolution probe call per inspected frame; its unique clip--stage selections and gallery embeddings are reusable.}
\label{tab:efficiency-breakdown}
\centering
\scriptsize
\textbf{(a) Mean online stage cost per query: busy seconds / model calls}\\[-1pt]
\begin{tabular}{@{}lrrrrrrrr@{}}
\toprule
Dataset & \multicolumn{2}{c}{Decomposition} & \multicolumn{2}{c}{Embedding} &
\multicolumn{2}{c}{Reranking} & \multicolumn{2}{c}{Verification} \\
\cmidrule(lr){2-3}\cmidrule(lr){4-5}\cmidrule(lr){6-7}\cmidrule(l){8-9}
& Time & Calls & Time & Calls & Time & Calls & Time & Calls \\
\midrule
WebVid-CoVR       & 15.603 & 0.996 & 1.434 & 1.000 & 47.898 & 74.015 & 4.815 & 0.253 \\
Dense-WebVid-CoVR &  3.733 & 0.261 & 1.437 & 1.000 & 31.848 & 49.160 & 1.613 & 0.093 \\
CoVR-R            &  1.897 & 0.111 & 1.483 & 1.000 & 13.741 & 22.075 & 0.721 & 0.041 \\
\bottomrule
\end{tabular}

\vspace{4pt}
\textbf{(b) Reusable preparation}\\[-1pt]
\begin{tabular}{@{}lrrrrrr@{}}
\toprule
Dataset & Selection tasks & Mean s/task & Calls/task & Total probe calls & Gallery s/item & Gallery calls \\
\midrule
WebVid-CoVR       & 3{,}958 & 2.824 & 33.916 & 134{,}240 & 1.420 & 1{,}000 \\
Dense-WebVid-CoVR & 3{,}958 & 2.634 & 33.916 & 134{,}240 & 1.422 & 1{,}000 \\
CoVR-R            & 3{,}954 & 2.889 & 33.918 & 134{,}112 & 1.425 & 1{,}000 \\
\bottomrule
\end{tabular}

\end{table*}

We profile the selected configuration on size-matched 1,000-query, 1,000-target cohorts from WebVid-CoVR, Dense-WebVid-CoVR, and the WebVid subset of the official CoVR-R annotation. Each run uses a target-video gallery with source self-masking, 16 workers, and identical model, frame, and routing settings; the generative seeds are fixed to 42 only for this controlled profile. Persistent embedding, prediction, and frame-bundle caches are disabled, so earlier runs cannot reduce the measured cost, while reuse within each invocation is retained to reflect normal batched execution.

Table~\ref{tab:efficiency} isolates the online query path: decomposition, composed-query embedding, reranking, and verification. Gallery encoding and query-independent temporal probing are excluded from this boundary and reported separately in Table~\ref{tab:efficiency-breakdown}. Mean online busy time is 69.751, 38.630, and 17.842 seconds per query on WebVid-CoVR, Dense-WebVid-CoVR, and CoVR-R, respectively. The corresponding compute is 76.264, 50.514, and 23.227 model inferences per query. A gate skip counts as zero calls; the totals therefore use the 996, 261, and 111 descriptions actually generated in the three runs and exclude no-op stage entries.

The cost difference follows the edit distribution rather than a configuration change. WebVid-CoVR activates expansion, decomposition, and verification for 75.4\%, 99.6\%, and 16.1\% of queries; the corresponding rates are 44.2\%, 26.1\%, and 6.5\% on Dense-WebVid-CoVR and 17.4\%, 11.1\%, and 3.6\% on CoVR-R. Thus concise, underspecified WebVid edits traverse the deepest path most often, whereas the more explicit CoVR-R edits usually terminate after a shallow candidate pass.

Table~\ref{tab:efficiency-breakdown} shows that repeated candidate scoring dominates online work: the reranker averages 47.898, 31.848, and 13.741 busy seconds per query across the same three datasets. Query embedding remains nearly constant at 1.434--1.483 seconds and one call per query, while the conditional generative stages scale with their activation rates. The reusable selector performs about 33.9 low-resolution frame-probe inferences per clip--stage selection, for 134,240, 134,240, and 134,112 calls across the three cohorts; gallery encoding adds one call per target. Target-side selections and gallery vectors can be amortized across future queries, while a new source still requires its own query-independent selection.

The evaluation-routine wall times are 5,330.73, 3,233.29, and 2,012.57 seconds, while the online-model interval unions are 4,416.26, 2,488.34, and 1,204.25 seconds. These wall times reflect the measured serving environment, overlap among 16 workers, and any request retries; they are neither hardware-independent latency nor a cross-system speed comparison. Model-call counts provide the complementary interface-level workload measure, but should not be interpreted as FLOPs or monetary cost. Table~\ref{tab:final-routing-workload} separately reports routing rates on the three complete evaluations.

\begin{table*}[!t]
\caption{Routing rates for the three complete evaluations, including embedding-gap skips, expansion and rescue rates, and mean expansion depth; these measurements expose the different workloads induced by each edit distribution.}
\label{tab:final-routing-workload}
\centering
\scriptsize
\begin{tabular}{@{}lrrrr@{}}
\toprule
Dataset & Gap skips & Expanded & Rescue rate & Mean rounds \\
\midrule
WebVid-CoVR & 2.86\% & 80.75\% & 14.51\% & 8.485 \\
Dense-WebVid-CoVR & 3.17\% & 46.50\% & 9.78\% & 8.000 \\
CoVR-R & 3.00\% & 27.11\% & 15.07\% & 6.109 \\
\bottomrule
\end{tabular}

\end{table*}

\begin{table*}[!t]
\ContinuedFloat
\caption[]{Final routing rates (continued: decomposition work).}
\label{tab:final-routing-workload-decomposition}
\centering
\scriptsize
\begin{tabular}{@{}lrrrr@{}}
\toprule
Dataset & Decomposed & Concise-edit route & Ambiguity route & Bypassed \\
\midrule
WebVid-CoVR & 99.80\% & 99.22\% & 0.59\% & 0.20\% \\
Dense-WebVid-CoVR & 34.44\% & 3.25\% & 31.19\% & 65.56\% \\
CoVR-R & 15.45\% & 0.00\% & 15.45\% & 84.55\% \\
\bottomrule
\end{tabular}

\end{table*}

\begin{table*}[!t]
\ContinuedFloat
\caption[]{Final routing rates (continued: verification work).}
\label{tab:final-routing-workload-verification}
\centering
\scriptsize
\begin{tabular}{@{}lrrrrr@{}}
\toprule
Dataset & Verified & Mean depth & Repair rate & Regression rate & Net R@1 \\
\midrule
WebVid-CoVR & 24.73\% & 2.718 & 1.88\% & 0.55\% & +1.33 \\
Dense-WebVid-CoVR & 13.11\% & 2.755 & 1.06\% & 0.08\% & +0.98 \\
CoVR-R & 6.76\% & 2.573 & 0.27\% & 0.15\% & +0.11 \\
\bottomrule
\end{tabular}

\end{table*}

The routing distribution depends strongly on edit style. WebVid-CoVR sends nearly all concise edits through decomposition, Dense-WebVid-CoVR relies primarily on ambiguity-based routing, and CoVR-R uses that route exclusively. Verification remains net positive on every benchmark, while the much lower activation on CoVR-R illustrates how a stronger candidate order narrows downstream work.

\section{Exact Model Interfaces and Reproducibility}\label{app:interfaces}

\begin{table*}[!p]
\caption{Effective final configuration, grouped by reusable, always-on, conditional candidate, conditional generative, and execution-boundary work. It records model roles, interfaces, frame policies, routing, generation, retries, concurrency, and reuse. The 15-frame setting is an upper bound per video input; short or unreadable clips can yield fewer, verification receives separate source and candidate inputs, and the selector may inspect additional low-resolution probes.}
\label{tab:full-configuration}
\centering
\tiny
\setlength{\tabcolsep}{3.2pt}
\begin{tabularx}{\textwidth}{@{}p{.12\textwidth}p{.14\textwidth}p{.13\textwidth}X p{.18\textwidth}@{}}
\toprule
Class & Component & Model & Input $\rightarrow$ output & Frames / pixels / time \\
\midrule
Reusable /\allowbreak{} offline & Gallery preparation & Deterministic data path & Validated target-only gallery $\rightarrow$ Deduplicated target list and query-independent frame selections & Uniform; up to 15 target frames; 512 px long side; timestamps off \\
Reusable for targets; online for a new source & Dynamic novelty probe & Qwen3-VL-Embedding-2B & Individual query-independent probe frames $\rightarrow$ L2-normalized per-frame vectors and adjacent novelty & 8/\allowbreak{}4/\allowbreak{}2 FPS for \([0,10)\)/\allowbreak{}\([10,30)\)/\allowbreak{}\([30,\infty)\) s; inverse-CDF selection; novelty-mixing coefficient 1.0; per-clip output budget 15; 256 px probe frames; timestamps off during probing \\
Always-on online & Composed Query Embedding & Qwen3-VL-Embedding-2B & Source video + raw modification text $\rightarrow$ Dense query vector, L2-normalized & Uniform; up to 15 source frames; 512 px long side; timestamps off \\
Reusable /\allowbreak{} offline & Reusable Gallery Encoding & Qwen3-VL-Embedding-2B & Target video only $\rightarrow$ One dense vector per target, L2-normalized & Uniform; up to 15 target frames; 512 px long side; timestamps off \\
Always-on online & Coarse retrieval and masking & Cosine similarity over normalized vectors & Query vector + reusable target matrix $\rightarrow$ Initial top five candidates & No additional video input; n/\allowbreak{}a; timestamps n/\allowbreak{}a \\
Conditional online candidate work & Confidence-Gated Adaptive Reranking & Qwen3-VL-Reranker-8B & Raw edit or one decomposed target description + candidate video; source video absent $\rightarrow$ Hosted continuous relevance score & Dynamic; up to 15 candidate frames; 256 px long side; timestamps off \\
Conditional online generative work & Conditional Query Decomposition & Qwen3.5-9B (shared generation model) & Source video + raw modification text $\rightarrow$ Generated description; whitespace trimmed and one enclosing quote pair removed (prompt requests one or two sentences) & Dynamic; up to 15 source frames; 256 px long side; timestamps on and expressed in seconds to two decimal places \\
Conditional online generative work & Confidence-Gated Candidate Verification & Qwen3.5-9B (shared generation model) & Source video + raw modification + one candidate video $\rightarrow$ Boolean or unclear verdict from a leading yes/\allowbreak{}no parser; strictness is prompt-defined & Dynamic; up to 15 source and 15 candidate frames; 512 px long side; timestamps on for both clips and expressed in seconds to two decimal places \\
Execution boundary & Concurrency and presampling & 16 evaluation workers; 16 dynamic-probe workers & Full active source/\allowbreak{}gallery task set $\rightarrow$ Presampled indices followed by lazy final frame bundles & Frame indices are presampled; complete frame bundles are not preloaded; stage-specific caps appear above; timestamps are rendered only during final decomposer/\allowbreak{}verifier bundle construction \\
Execution boundary & Timeouts and retry layers & All hosted model clients & Stateless model requests and failed evaluator items $\rightarrow$ Bounded transport recovery followed by role-specific failed-item replay & Probe requests use their separate dynamic policy; n/\allowbreak{}a; timestamps n/\allowbreak{}a \\
\bottomrule
\end{tabularx}

\end{table*}

\begin{table*}[!p]
\ContinuedFloat
\caption[]{Effective final configuration (continued: routing, execution, and caching).}
\label{tab:full-configuration-routing}
\centering
\tiny
\setlength{\tabcolsep}{3.2pt}
\begin{tabularx}{\textwidth}{@{}p{.16\textwidth}X p{.20\textwidth}p{.25\textwidth}@{}}
\toprule
Component & Routing / thresholds & Sampling / retries & Cache and reuse \\
\midrule
Gallery preparation & Source self-mask on; positive limit applied before gallery rebuild & Deterministic & Selections and one target embedding are reusable across queries \\
Dynamic novelty probe & No text/\allowbreak{}query conditioning; uniform fallback if novelty mass is zero & 16 concurrent probe workers in final runs & Probe vectors key clip/\allowbreak{}model/\allowbreak{}FPS/\allowbreak{}resolution/\allowbreak{}version; selections also key budget/\allowbreak{}ratio/\allowbreak{}bands/\allowbreak{}rates \\
Composed Query Embedding & Joint input; role-specific instruction & One query embedding request & In-run and atomic cross-run embedding cache keyed by role, content, instruction, model, and resolved frames \\
Reusable Gallery Encoding & An explicit empty target instruction is transmitted & One embedding request per distinct target & Stored normalized gallery matrix is reused by every query \\
Coarse retrieval and masking & Mask source item when present; skip reranking only if the gap between the two highest embedding scores exceeds 0.25 & Equality at 0.25 reranks & Gallery matrix reused; search is query-specific \\
Confidence-Gated Adaptive Reranking & Initial 5; append 5 while the highest reranker score is below 0.70; at most 9 expansions /\allowbreak{} 50 candidates; expose top 10 downstream & No client instruction is supplied; unchanged pair scores are reused & Prediction key includes model, query form/\allowbreak{}content, frame signature, resolution, timestamps, and prompt \\
Conditional Query Decomposition & Early if ASCII word count is below 10; otherwise late if the highest reranker score is at least 0.25 and its relative gap to the second-highest score is below 0.50; reranker only & Thinking 2048; max 2560 tokens; temperature unset -\textgreater{} 0.6, top-p 0.95, top-k 20; 10 failed-item follow-up passes after transport handling; unseeded & One description is reused by reranking; failure/\allowbreak{}empty output falls back to raw text \\
Confidence-Gated Candidate Verification & Require a highest reranker score of at least 0.15 and at least two candidates; scan the contiguous top-10 prefix whose relative score decrease from the leader is below 0.25; stop at first yes & Thinking 2048; max 2560 tokens; temperature unset -\textgreater{} 0.6, top-p 0.95, top-k 20; 10 failed-item follow-up passes after transport handling; unseeded & Stable promotion; skip/\allowbreak{}failure/\allowbreak{}all-no retains prior valid order \\
Concurrency and presampling & Queries/\allowbreak{}items parallel; expansion pages and verifier scans sequential within a query & Early decomposition overlaps query then gallery embedding after the prepass & Presampling resolves probes/\allowbreak{}indices only; decode, resize, overlay, and JPEG packing remain lazy \\
Timeouts and retry layers & General calls: 300 s timeout, 5 transport retries; dynamic probes: 120 s timeout, 5 transport retries & Transport retries 429/\allowbreak{}500/\allowbreak{}502/\allowbreak{}503/\allowbreak{}504 and connection/\allowbreak{}read failures; backoff factor 0.5 s, cap 30 s, jitter 0.5 s; general pool 64, probe pool 16; decomposer/\allowbreak{}verifier add up to 10 failed-item passes with 2 s waits & Transport retry repeats one stateless request; failed-item passes rerun only unresolved evaluator items \\
\bottomrule
\end{tabularx}

\end{table*}

The retrieval encoder is Qwen3-VL-Embedding-2B. Its final query form is source video plus raw edit under a target-oriented retrieval instruction; target videos use an explicit neutral instruction. The same embedding family supplies the language-free low-resolution temporal probes. Qwen3-VL-Reranker-8B receives raw or decomposed text with each candidate video and returns a continuous relevance score. Decomposition and verification share Qwen3.5-9B but retain separate prompts, request layouts, output parsers, resolutions, and reusable prediction records.

\begin{table*}[!p]
\caption{Exact model-interface text and request contracts. Embedding rows preserve the literal role instructions, the reranker row records the tracked serving-template fields and unset client instruction, and the generative rows preserve the complete runtime system prompts, labeled user-message order, output parsing, timestamp note, effective sampling parameters, thinking budget, maximum generation length, seed policy, and both transport- and evaluator-level retry layers.}
\label{tab:prompt-interfaces}
\centering
\tiny
\setlength{\tabcolsep}{3.2pt}
\begin{tabularx}{\textwidth}{@{}P{.16\textwidth}P{.16\textwidth}P{.21\textwidth}P{.17\textwidth}Y@{}}
\toprule
Role & Model & User layout & Output & Timestamp / runtime \\
\midrule
Composed query embedding & Qwen3-VL-Embedding-2B & source video; raw modification text & one dense vector & timestamps off; L2-normalize before cosine search \\
Source-video-only embedding ablation & Qwen3-VL-Embedding-2B & source video & one dense vector & timestamps off; Ablation interface; not the final composed-query form \\
Text-only embedding ablation & Qwen3-VL-Embedding-2B & raw modification text & one dense vector & timestamps n/\allowbreak{}a; Ablation interface; not the final composed-query form \\
Target gallery embedding & Qwen3-VL-Embedding-2B & target video only; an explicit empty instruction is transmitted & one dense vector & timestamps off; L2-normalize once and store \\
Query-independent dynamic-frame probe & Qwen3-VL-Embedding-2B & one probe image; empty system instruction is transmitted & one dense vector per probed frame; adjacent 1-cos supplies novelty & timestamps off; query independent; normalized frame vectors; no language query \\
Candidate reranking & Qwen3-VL-Reranker-8B & raw edit or decomposed target description as the search request; candidate video as the document; source video absent; no client instruction is supplied & hosted continuous relevance score consumed directly & timestamps off; client does not reconstruct yes/\allowbreak{}no token probabilities \\
Conditional query decomposition & Qwen3.5-9B & source-video frames followed by the raw modification text & prompt requests one or two plain sentences; parser trims whitespace and removes one enclosing matching quote pair only & timestamps on; timestamp note appended; thinking on; budget 2048; max 2560; effective temperature/\allowbreak{}top-p/\allowbreak{}top-k 0.6/\allowbreak{}0.95/\allowbreak{}20; 300 s request timeout + 5 transport retries; up to 10 failed-item follow-up passes with 2 s waits; unseeded \\
Strict candidate verification & Qwen3.5-9B & source-video frames; raw modification text; candidate-video frames & prompt requests one word; parser trims/\allowbreak{}lowercases, strips leading quote/\allowbreak{}backtick/\allowbreak{}asterisk/\allowbreak{}space characters, accepts a response beginning yes or no, and otherwise returns unclear & timestamps on for both videos; timestamp note appended; thinking on; budget 2048; max 2560; effective temperature/\allowbreak{}top-p/\allowbreak{}top-k 0.6/\allowbreak{}0.95/\allowbreak{}20; 300 s request timeout + 5 transport retries; up to 10 failed-item follow-up passes with 2 s waits; unseeded \\
\bottomrule
\end{tabularx}

\end{table*}

\begin{table*}[!p]
\ContinuedFloat
\caption[]{Exact model-interface contracts (continued: embedding instructions).}
\label{tab:prompt-interfaces-embedding}
\centering
\scriptsize
\setlength{\tabcolsep}{3.2pt}
\begin{tabularx}{\textwidth}{@{}p{.25\textwidth}X@{}}
\toprule
Embedding role & Exact instruction \\
\midrule
Composed query embedding & Given this source video with the desired modification text, represent a potential target video for retrieval. \\ 
Source-video-only embedding ablation & Represent this video for retrieval. \\ 
Text-only embedding ablation & Represent this text query for retrieving the described video. \\ 
Target gallery embedding & \textless{}empty string transmitted\textgreater{} \\ 
Query-independent dynamic-frame probe & \textless{}empty string transmitted\textgreater{} \\ 
\bottomrule
\end{tabularx}

\end{table*}

\begin{table*}[!p]
\ContinuedFloat
\caption[]{Exact model-interface contracts (continued: reranker serving template).}
\label{tab:prompt-interfaces-reranker}
\centering
\tiny
\setlength{\tabcolsep}{3.2pt}
\textbf{Candidate reranking} \quad Qwen3-VL-Reranker-8B
\par\smallskip
\noindent\textbf{Prompt contract.}
The system asks whether the candidate document satisfies the search request under the supplied retrieval instruction, and restricts the verbal judgment to ``yes'' or ``no.'' The user turn contains three semantic fields: a retrieval instruction, the search request, and the candidate document. A supplied system instruction fills the first field; otherwise the service uses the default instruction ``Given a search query, retrieve relevant candidates that answer the query.'' The search request is the raw edit or decomposed target description, and the candidate document is the candidate video; the source video is absent. No client instruction is supplied in our configuration, so the default retrieval instruction applies.
\par\smallskip\noindent\textbf{Output:} The hosted continuous relevance score is consumed directly; the client does not reconstruct token probabilities from the verbal judgment.

\end{table*}

\begin{table*}[!p]
\ContinuedFloat
\caption[]{Exact model-interface contracts (continued: decomposition prompt).}
\label{tab:prompt-interfaces-decomposition}
\centering
\tiny
\setlength{\tabcolsep}{3.2pt}
\textbf{Conditional query decomposition} \quad Qwen3.5-9B
\par\smallskip
\begin{Verbatim}[fontsize=\tiny,breaklines=true,breakanywhere=true]
You are a professional video analyst helping with composed video retrieval.
You are given a query video and a modification text describing the changes to apply to it.
Your task is to write a single, self-contained description of the TARGET video: the video obtained by applying the described modifications to the query video.
Keep every detail of the query video that the modification does not change (subjects, actions, scene, setting, mood), and apply every requested change (a modification may alter the subject, the action, or the context, and may bundle several changes at once).
Write one or two plain sentences describing the target video as it would actually appear. Do not mention the query video, the modification, or the retrieval task; output only the description.

A timestamp is drawn in each frame for reference only and is not part of the video. Videos do not have to be of the same length.
\end{Verbatim}
\smallskip\noindent\textbf{User content:} Query video: \textless{}source frames\textgreater{}; Modification text: \textless{}raw edit\textgreater{}
\par\noindent\textbf{Output:} prompt requests one or two plain sentences; parser trims whitespace and removes one enclosing matching quote pair only

\end{table*}

\begin{table*}[!p]
\ContinuedFloat
\caption[]{Exact model-interface contracts (continued: verification prompt).}
\label{tab:prompt-interfaces-verification}
\centering
\tiny
\setlength{\tabcolsep}{3.2pt}
\textbf{Strict candidate verification} \quad Qwen3.5-9B
\par\smallskip
\begin{Verbatim}[fontsize=\tiny,breaklines=true,breakanywhere=true]
You are a professional video analyst.
You are given a query video with a modification text mentioning required changes to the query video, along with a potential target video.
First, analyze the query video with modification text, and the target video. Next, answer whether the target video is relevant to user's query.
Answer 'yes' only if everything mentioned in the modification query is clearly presented in the target video, without ambiguities.It can still be 'yes' if the found differences are not specifically mentioned in the modification query.Answer 'no' if at least one thing mentioned in the modification query is missing, only weakly implied, contradicted, or you are unsure.
Only answer 'yes' or 'no' with one word.

A timestamp is drawn in each frame for reference only and is not part of the video. Videos do not have to be of the same length.
\end{Verbatim}
\smallskip\noindent\textbf{User content:} Query video: \textless{}source frames\textgreater{}; Modification text: \textless{}raw edit\textgreater{}; Target video: \textless{}candidate frames\textgreater{}
\par\noindent\textbf{Output:} prompt requests one word; parser trims/\allowbreak{}lowercases, strips leading quote/\allowbreak{}backtick/\allowbreak{}asterisk/\allowbreak{}space characters, accepts a response beginning yes or no, and otherwise returns unclear

\end{table*}

Both generative roles use an enabled reasoning mode with a 2,048-token reasoning budget and a 2,560-token maximum response. With temperature unset, the effective defaults are temperature 0.6, top-\(p\) 0.95, and top-\(k\) 20; final accuracy evaluations are unseeded. General requests use a 300-second timeout, bounded transport retries with exponential backoff and jitter, and a shared connection pool. Temporal probes use a shorter timeout and a separate pool. After transport handling, unresolved decomposition or verification items receive bounded evaluator-level retries. The decomposition parser trims whitespace and one enclosing quote pair; the verifier normalizes leading formatting and accepts responses beginning with yes or no.

\subsection{Caching, Presampling, Concurrency, and Reproducibility}\label{app:reproducibility}

Four reuse families are independent. Embeddings use in-memory reuse plus atomic binary vector entries keyed by model, instruction, role, content, resolved frames, and pixel policy. Reranker, decomposer, and verifier predictions use structured records keyed by stage, model, prompt, query content, frame signature, resolution, timestamps, reasoning, and sampling. Temporal-probe vectors additionally key clip, probe model, frame rate, resolution, and selector version; selected-index records include the output budget, novelty mixture, duration bands, and rates. Final frame bundles use a bounded in-memory least-recently-used store and optional compressed structured storage keyed by clip, exact indices, pixel cap, timestamp state, image-encoding policy, and version.

Writes are atomic and malformed entries become misses. Normal accuracy evaluation permits cross-invocation reuse, whereas timed evaluations redirect or disable the relevant persistent stores while retaining safe reuse within the invocation. This prevents previous evaluations from shrinking the measured cold boundary without duplicating work inside one evaluation.

The final configuration resolves temporal probes and selected indices before inference but does not preload complete frame bundles. Final decoding, resizing, timestamp rendering, and image packing remain lazy. Candidate frames are reused across repeated candidate calls, but every uncached query-conditioned reranker or verifier pair remains online work. Independent items and queries run concurrently; expansion rounds and verifier scans remain sequential within a query, and early decomposition can overlap embedding before joining at reranking.

\clearpage
\section{Limitations and Future Directions}\label{app:limitations}

The study is comprehensive stage-wise but not exhaustive over every foundation-model family, and stochastic generative ablations are primarily single evaluations. Dynamic selection is query-independent and can under-allocate slowly evolving evidence; training-free query-aware allocation is a natural next step. Fixed gates and the bounded candidate ceiling calibrate well on the evaluated galleries but remain a hard recall boundary because targets omitted upstream are unreachable by verification. Target-only galleries and single-positive annotations also under-measure multiple-valid-answer uncertainty, as the campfire case demonstrates. Strict verification is net positive in every final evaluation but can still reject a correct near-tied candidate. Future work should examine calibration under domain shift, longer and streaming video, multiple-valid-target evaluation, alternative interface-compatible model families, query-aware training-free selection, and adaptive candidate budgets that preserve the cascade's reusable/conditional separation.

\end{document}